\documentclass{article}
\usepackage{amsmath,amsfonts,bm}

\def\eqref#1{equation~\ref{#1}}
\def\1{\bm{1}}

\DeclareMathAlphabet{\mathsfit}{\encodingdefault}{\sfdefault}{m}{sl}
\SetMathAlphabet{\mathsfit}{bold}{\encodingdefault}{\sfdefault}{bx}{n}

\DeclareMathOperator*{\argmax}{arg\,max}

\usepackage{microtype}
\usepackage{graphicx}
\usepackage{booktabs} % for professional tables
\PassOptionsToPackage{dvipsnames}{xcolor}

\usepackage{hyperref}

\usepackage[accepted]{icml2025}

\usepackage{amsmath}
\usepackage{amssymb}
\usepackage{mathtools}
\usepackage{amsthm}

\usepackage[capitalize,noabbrev]{cleveref}

\theoremstyle{plain}

\theoremstyle{definition}

\theoremstyle{remark}

\usepackage[textsize=tiny]{todonotes}

\usepackage{graphicx}
\usepackage{color}
\usepackage[utf8]{inputenc}

\renewcommand{\algorithmiccomment}[1]{\hfill\textcolor{gray}{// #1}}

\usepackage{setspace}

\usepackage{float}
\usepackage{amsmath}
\usepackage{commath}
\usepackage{flexisym}
\usepackage{relsize}\usepackage{amssymb}
\usepackage{wrapfig}

\definecolor{BrickRed}{rgb}{0.6,0,0}
\definecolor{RoyalBlue}{rgb}{0,0,0.8}
\definecolor{Tdgreen}{rgb}{0,0.4,0.7}
\hypersetup{linkcolor=BrickRed}

\usepackage{subcaption} % for horizontal subfigure
\usepackage{bbold} %for math symbol 1
\usepackage{tikz}
\usepackage[normalem]{ulem}
\useunder{\uline}{\ul}{}
\newcommand{\rebuttal}[1]{#1}

\newcommand{\std}{0.7}

\newcommand{\system}{{TTA with binary feedback}}
\newcommand{\method}{BiTTA}

\usepackage{hyperref}
\usepackage{url}

\usepackage[utf8]{inputenc} % allow utf-8 input
\usepackage[T1]{fontenc}    % use 8-bit T1 fonts
\hypersetup{colorlinks}
\usepackage{booktabs}       % professional-quality tables
\usepackage{amsfonts}       % blackboard math symbols
\usepackage{nicefrac}       % compact symbols for 1/2, etc.
\usepackage{makecell}       % used in tables
\usepackage{multicol} 
\usepackage{multirow}
\usepackage{xspace}
\usepackage{multirow}
\usepackage{adjustbox}
\usepackage{colortbl}
\usepackage{duckuments}
\usepackage{lipsum}
\usepackage{enumitem}
\usepackage{kotex}
\usepackage{amsmath}
\usepackage{graphicx}
\usepackage{tabularx}
\usepackage{ragged2e}
\usepackage{amsmath}
\usepackage{amsthm}

\newcolumntype{Y}{>{\centering\arraybackslash}X}
\newcolumntype{Z}{>{\raggedRight\arraybackslash}X}
\newcolumntype{R}[1]{>{\RaggedLeft\arraybackslash}p{#1}}

\usepackage{rotating}

\definecolor{Gray}{gray}{0.9}

\begin{document}

\twocolumn[
\icmltitle{Test-Time Adaptation with Binary Feedback}



\icmlsetsymbol{equal}{*}

\begin{icmlauthorlist}
\icmlauthor{Taeckyung Lee}{KAIST}
\icmlauthor{Sorn Chottananurak}{KAIST}
\icmlauthor{Junsu Kim}{KAIST}
\icmlauthor{Jinwoo Shin}{KAIST}
\icmlauthor{Taesik Gong}{UNIST,equal}
\icmlauthor{Sung-Ju Lee}{KAIST,equal}
% \icmlauthor{Firstname1 Lastname1}{equal,yyy}
% \icmlauthor{Firstname2 Lastname2}{equal,yyy,comp}
% \icmlauthor{Firstname3 Lastname3}{comp}
% \icmlauthor{Anonymous Author(s)}{yyy}
\end{icmlauthorlist}

\icmlaffiliation{KAIST}{KAIST}
\icmlaffiliation{UNIST}{UNIST}

% \icmlcorrespondingauthor{Anonymous Author(s)}{first1.last1@xxx.edu}
% \icmlcorrespondingauthor{Taeckyung Lee}{taeckyung@kaist.ac.kr}
\icmlcorrespondingauthor{Taesik Gong}{taesik.gong@unist.ac.kr}
\icmlcorrespondingauthor{Sung-Ju Lee}{profsj@kaist.ac.kr}

\icmlkeywords{Machine Learning, ICML, Test-Time Adaptation, Domain Adaptation, Online Learning}

\vskip 0.3in
]



\printAffiliationsAndNotice{${}^*$Corresponding authors.} % otherwise use the standard text.

% non-interest -> Irrelevant -> non-interest?

\begin{abstract}

Deep learning models perform poorly when domain shifts exist between training and test data. Test-time adaptation (TTA) is a paradigm to mitigate this issue by adapting pre-trained models using only unlabeled test samples. 
However, existing TTA methods can fail under severe domain shifts, while recent active TTA approaches requiring full-class labels are impractical due to high labeling costs.
To address this issue, we introduce a new setting of TTA with binary feedback. This setting uses a few binary feedback inputs from annotators to indicate whether model predictions are correct, thereby significantly reducing the labeling burden of annotators. 
Under the setting, we propose \method{}, a novel dual-path optimization framework that leverages reinforcement learning to balance binary feedback-guided adaptation on uncertain samples with agreement-based self-adaptation on confident predictions. 
Experiments show \method{} achieves 13.3\%p accuracy improvements over state-of-the-art baselines, demonstrating its effectiveness in handling severe distribution shifts with minimal labeling effort.
The source code is available at \url{https://github.com/taeckyung/BiTTA}.

\end{abstract}

\section{Introduction}\label{sec:intro}

Deep learning has revolutionized various fields, including computer vision~\cite{imagenet}, speech recognition~\cite{gulati2020conformer}, and natural language processing~\cite{brown2020language}. However, deep models often suffer from domain shifts, where discrepancies between training and test data distributions lead to significant performance degradation. For example, autonomous driving systems might struggle with new types of vehicles or unexpected weather conditions that differ from the training data~\cite{sakaridis2018semantic}.
% image classification, natural language processing, and object detection~\cite{7780459, ronneberger2015u, 10.1145/3422622, vaswani2017attention, brown2020language}. However, in real-world applications, deep learning models often encounter domain shifts, where the training data distribution differs from the test data distribution, leading to significant drops in performance. For example, object detection models might struggle with previously unseen variations of objects or changes in environmental conditions such as weather.

\begin{figure*}[t]
    \centering
    \begin{minipage}[t]{0.7\textwidth}

        \centering
        \includegraphics[width=\linewidth]{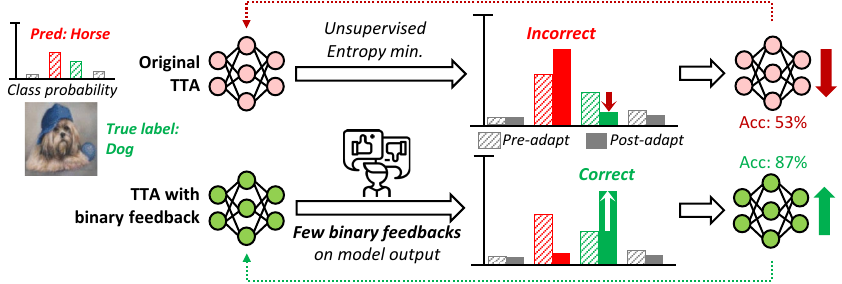}
        \caption{Overview of TTA with binary feedback. Traditional TTA algorithms often fail under severe distribution shifts due to the risk of unlabeled-only adaptation. Our proposed \system{} addresses this challenge by offering a few binary feedback (\textit{correct} or \textit{incorrect}) on selected model predictions. TTA with binary feedback significantly improves the adaptation performance with minimal labeling effort, enabling a practical and scalable TTA paradigm for real-world applications.
        % thereby potentially maintaining true label confidence.
        }~\label{fig:scenario}

    \end{minipage}
    \hspace{0.05cm}
    \begin{minipage}[t]{0.28\textwidth}

        \centering
        \includegraphics[width=\linewidth]{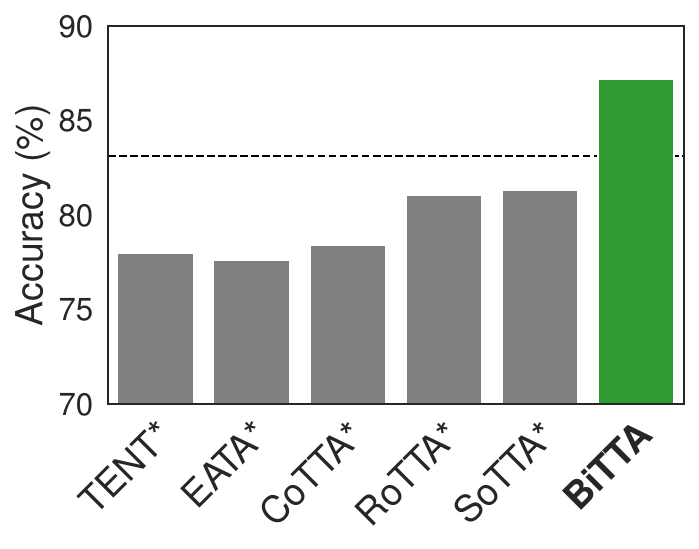}
        \caption{Accuracy (\%) of TTA methods with binary feedback on CIFAR10-C. The asterisk indicates a modified algorithm to utilize binary feedback. The dotted line is full-class active TTA (SimATTA).}\label{fig:performance}
        
    \end{minipage}

\end{figure*}

Test-time adaptation (TTA)~\cite{tent} dynamically adapts the pre-trained models in real time using only unlabeled test samples. Hence, it is a viable solution to domain shifts.  
However, without ground-truth labels, existing TTA methods are vulnerable to adaptation failures or suboptimal accuracies in severe distribution shifts~\cite{note, sar, sotta, aetta, rdumb}. This is largely due to relying on self-supervised metrics such as entropy or confidence~\cite{tent, sotta} that could be unreliable when adaptation fails~\cite{aetta}.

This limitation motivates the potential value of incorporating human feedback into the adaptation loop.
% Recent advances in language models have demonstrated the transformative power of human feedback in model adaptation. 
In language models, reinforcement learning from human feedback (RLHF)~\cite{ouyang2022training} has emphasized the critical role of human feedback in aligning large language models with human intent. 
% \rebuttal{However, existing TTA methods typically lack such direct human interaction, relying instead on self-supervised metrics such as entropy or confidence~\cite{tent, sotta}, which can be unreliable under severe distribution shifts. For example, such metrics remain at high values even with the model failures~\cite{aetta}}.
The key challenge in TTA is making human feedback practical and scalable. 
% Querying a full labeling of test samples is prohibitively expensive for end-users~\cite{joshi2010breaking}, necessitating a lightweight labeling approach for TTA.
While full labeling of test samples is prohibitively expensive~\cite{joshi2010breaking}, recent studies on language models have shown that binary feedback (e.g., thumbs up/down) can effectively guide model behaviors~\cite{shankar2024validates}.

% Inspired by recent studies on large language models that demonstrate how lightweight binary feedback (e.g., thumbs up/down) can effectively guide model behavior~\cite{shankar2024validates},
% Motivated by the fact that adaptation models already provide a guess for each test sample,
Inspired by this,
we introduce the \textbf{TTA with binary feedback} setting (Figure~\ref{fig:scenario}), which uses binary feedback on the model prediction (\textit{correct} or \textit{incorrect}) to guide adaptation while maintaining efficiency. 
% This aligns with recent efforts to adjust the language model behavior with thumbs up/down signals~\cite{shankar2024validates}.
Our binary-feedback approach requires only minimal label information, significantly reducing labeling costs compared with full-class active TTA~\cite{atta}, mitigating the interaction bottlenecks to enable real-world applications.

\begin{figure*}[t]
    \centering
    \includegraphics[width=\linewidth]{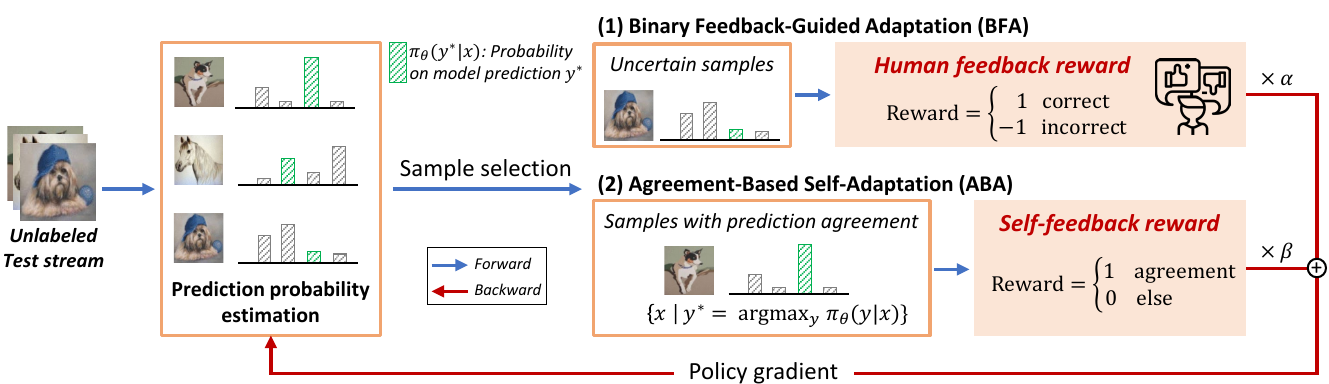}
    \caption{Overview of \method{} algorithm. \method{} implements a reinforcement learning-based dual-path optimization that estimates prediction probabilities using MC-dropout. It computes policy gradients from two complementary signals: (1) Binary Feedback-guided Adaptation (BFA) on uncertain samples, using binary rewards of $\pm 1$, and (2) Agreement-Based self-Adaptation (ABA) on confident, unlabeled samples, using reward 1. By jointly optimizing both paths, \method{} enables robust adaptation under dynamic distribution shift scenarios.
    }
    \label{fig:overview}
    % \vspace{-0.3cm}
    % \hfill
\end{figure*}

As a solution, we propose \textbf{\method{}}, a dual-path optimization framework for \system{} that incorporates both binary feedback and unlabeled samples. Motivated by the recent reinforcement learning studies that show effectiveness in incorporating human feedback in the optimization process~\cite{ouyang2022training, fan2024reinforcement, black2024training}, 
\method{} leverages reinforcement learning to effectively balance two complementary adaptation strategies (Figure~\ref{fig:overview}): \textit{Binary Feedback-guided Adaptation (BFA)} on uncertain samples and \textit{Agreement-Based self-Adaptation (ABA)} on confident samples. Using Monte Carlo dropout~\cite{dropout} for policy estimation and uncertainty assessment, we select uncertain samples for binary feedback in BFA while utilizing samples with high prediction agreement in ABA. This dual approach enables \method{} to adapt to new uncertain patterns (via BFA) while maintaining confidence in correct predictions (via ABA), achieving robust performance improvements.

We evaluate \method{} under the \system{} setting with various test-time distribution shift scenarios, including three image corruption datasets (CIFAR10-C, CIFAR100-C, and Tiny-ImageNet-C) and two domain generalization scenarios (domain-wise and mixed data streams). 
Comparisons with TTA and active TTA methods demonstrate that \method{} achieves an accuracy improvement of 13.3\%p on average. 
% Notably, \method{} is the only method to outperform full-labeled active TTA (Figure~\ref{fig:performance}), and 
\rebuttal{Moreover, \method{} outperforms active TTA utilizing full-class feedback from the oracle (the ground-truth label, Figure~\ref{fig:performance}) and the state-of-the-art foundational model (GPT-4o, Figure~\ref{fig:acc_atta}), despite using only binary feedback.}
These results highlight the importance and effectiveness of \method{} and TTA with binary feedback, enabling robust adaptation with minimal labeling effort.

Our main contributions are as follows:
\vspace{-0.2cm} 
\begin{itemize}
    \item We propose a lightweight and scalable setting of TTA with binary feedback (Figure~\ref{fig:scenario}), offering \textit{correct} or \textit{incorrect} feedback on selected model predictions.
    \item We develop BiTTA, a dual-path optimization strategy combining two complementary signals from binary feedback-guided adaptation and agreement-based self-adaptation with a reinforcement learning framework.
    \item We perform extensive experiments that show BiTTA outperforms TTA and active TTA baselines by 13.3\%p. Comparisons with full-class active TTA indicate the effectiveness of BiTTA and TTA with binary feedback.
\end{itemize}
\section{\textls[-1]{Test-Time Adaptation with Binary Feedback}}\label{sec:background}

\rebuttal{We propose \system{}, a novel TTA setting for adapting pre-trained models during test time using binary feedback from an oracle, which indicates whether a model's predictions are correct~(Figure~\ref{fig:scenario}). 
% Unlike traditional test-time adaptation methods that operate without any feedback, or active approaches that require full class labels, \system{} only needs simple yes/no feedback about prediction correctness. 
% \system{} simply requires them to indicate whether the model's current prediction is correct or not (Figure~\ref{fig:scenario}), instead of the blind adaptation (traditional TTA) or asking ``what is the correct class among [\dots] classes'' (full-label active TTA). 
This real-time feedback seamlessly integrates into the adaptation process, enabling continuous refinement of the model's performance.}

\system{} addresses the critical challenge of adapting pre-trained models to domain shifts with minimal labeling effort. 
% Unlike methods that require full-class labels, \system{} leverages simple binary feedback to guide the adaptation process. 
% This reduces labeling costs and mitigates the interaction bottleneck, making it feasible for real-world applications where obtaining labeled data is impractical.
% BATTA offers two advantages over active full-labeling Test-Time Adaptation (TTA) approaches.
% (1) \textit{Reduced labeling cost}: Full labeling is as expensive as the maximum $O(N)$ trial of binary labeling;. %, demonstrating the lightweight mechanism. 
% (2) 
From an information-theory perspective, full-class labeling requires $ \log ({\tt num\_class})$ times more bits than binary feedback to encode the same information~\citep{mackay2003information}.
Empirical studies on human annotation also validate the efficiency of binary feedback: full-class labeling on 50-class takes 11.7 seconds on average with a 12.7\% error rate, whereas binary comparison requires only 1.6 seconds with 0.8\% error rate~\citep{joshi2010breaking}.

Although binary feedback provides only a single bit, the feedback is based on the adapted model's prediction, which is typically better than chance. This makes the feedback more informative and allows it to directly guide model behaviors.
Therefore, \system{} is an efficient and practical framework for real-world TTA applications.

% demonstrate that object comparison (e.g., comparing an image with a model prediction in BATTA) requires a lower labeling overhead than full-class labeling, making BATTA more efficient and practical for real-world applications.

% One might consider using large foundation models (e.g., GPT-4) as an oracle for providing feedback during test-time adaptation. However, our experiments in Appendix~\ref{app:ablation} show that even state-of-the-art foundation models perform worse than traditional TTA methods on distribution-shifted data. On the other hand, human feedback plays a crucial role in training/aligning foundational models~\cite{ouyang2022training}. This further highlights the unique value of human judgment in handling distribution shifts, necessitating our binary feedback mechanism.

% \vspace{-0.15cm}\paragraph{Feedback mechanism.} 
% In \system{}, an oracle provides a few binary feedbacks indicating whether the model's prediction is correct. This real-time feedback is integrated into the system, enabling continuous model adaptation. The binary feedback mechanism is illustrated in Figure~\ref{fig:scenario}, where the oracle evaluates the model's predictions, and the feedback memory is updated accordingly.

\vspace{-0.15cm}\paragraph{Notation.}
Let \( x \) denote a test sample selected for binary-feedback labeling, and \( y^* = \argmax_y f_\theta (y | x) \) is the class prediction output of the model parameterized by \( \theta \). The binary feedback \( B (x, y) \) is defined as:
\begin{equation}
B (x, y) = \begin{cases} 
1 & \text{if $y$ is correct,} \\
-1 & \text{if $y$ is incorrect.}
\end{cases}
\end{equation}
Accordingly, each binary-feedback sample is represented as the tuple $({x}, y^*, B(x, y^*))$, consisting of the input instance, the model's predicted label, and the binary feedback.
% with target instance ${x}$, model prediction label $y^*$, and binary feedback $B$.

\section{\method{}: Dual-path Optimization to Learn with Binary Feedback}

\paragraph{Motivation.} 
% Inspired by recent advances in reinforcement learning with human feedback (RLHF, \cite{ouyang2022training}) in large-language model training, 
Recent advancements in reinforcement learning from human feedback (RLHF)~\cite{ouyang2022training} have demonstrated the effectiveness of incorporating sparse feedback signals in large language model training. 
Inspired by this, we propose \method{}, a reinforcement learning (RL) based approach for \system{} that effectively adapts to distribution shifts using minimal labeling effort (Figure~\ref{fig:overview}). 
% Test-time adaptation (TTA) is crucial in real-world scenarios where deployed models encounter data distributions different from their training sets, potentially leading to performance degradation. However, acquiring full labels for adaptation is often impractical or expensive. 
\method{} leverages binary feedback as a reinforcement signal, offering several key advantages for TTA. 
(i)~Binary feedback can be seamlessly incorporated as non-differentiable rewards in the RL framework, enabling the model to learn from minimal supervision~\citep{zoph2017neural, yoon2020data}. (ii)~The RL framework allows for integrating binary feedback and unlabeled samples into a single objective function optimized through policy gradient methods. By combining sparse binary-feedback samples with unlabeled data, \method{} provides a robust framework with minimal labeling effort, making TTA more feasible for real-world applications.

% By formulating TTA as an RL problem, \method{} incorporates uncertainty estimation and sample selection strategies within a unified learning framework, leading to a more cohesive and theoretically grounded approach. This formulation enables \method{} a practical solution for maintaining model performance in dynamic real-world environments.

\vspace{-0.15cm}\paragraph{Policy gradient modeling.}
Given a batch of test samples $\mathcal{B} = \{x_1, \ldots, x_n\}$, our goal is to adapt the model parameters $\theta$ to improve performance on the test distribution. 
% A key challenge in \system{} is that binary feedback is inherently non-differentiable, making it difficult to optimize directly. To address this, w
We formulate the test-time adaptation process as an RL problem by assigning test-time input $x \sim \mathcal{B}$ as a state, the model prediction $y^* = f_\theta (x)$ as an action, and the corresponding prediction probability $\pi_\theta(y|x)$ as a policy, which objective is maximizing the expected reward, defined as:
\begin{equation}
    J(\theta) = \mathbb{E}_{x\sim\mathcal{B}, y\sim\pi_\theta(y|x)}[R(x, y)],
\end{equation}
% this is calculated on every test batch. 
where $R(x, y)$ represents the reward function defined later. This optimization is performed for each test batch, allowing continuous adaptation to the evolving test distribution. 
% entire test batch thereby calculating the policy gradient per batch instead of per-sample approach~\citep{zhao2023test}. This aligns with efficient adaptation in test time~\citep{tent, eata}.

As binary feedback is a non-differentiable function, we employ the \textsc{Reinforce} algorithm~\citep{williams1992simple}, also known as the ``log-derivative trick''. This method allows us to estimate the gradient of the expected reward with respect to the model parameters:
\begin{equation}
    \nabla_\theta J(\theta) = \mathbb{E}_{x\sim\mathcal{B}, y\sim\pi_\theta(y|x)}[R(x, y) \nabla_\theta \log \pi_\theta(y|x)].
\end{equation}
By using this gradient estimator, we can effectively optimize our model parameters using stochastic gradient ascent.

To estimate the policy $\pi_\theta$, we adopt Monte Carlo (MC) dropout~\citep{dropout}, a practical Bayesian approximation technique that enables robust uncertainty estimation without architectural changes. MC-dropout performs multiple stochastic forward passes with dropout enabled at test time, allowing the model to capture epistemic uncertainty in its predictions.

Formally, we approximate the policy $\pi_\theta(y|x)$ as the mean of softmax outputs across $N$ stochastic forward passes:
\begin{equation}
\label{eq:policy}
    \pi_\theta(y|x) = \frac{1}{N} \sum_{n=1}^{N} f_{\theta}^{d} (y | x),
\end{equation}
where $f_\theta^{d}$ denotes the model with dropout applied during inference, and $N$ is the number of samples drawn.

This approach is crucial for test-time adaptation under distribution shift. Unlike standard softmax outputs, which are often miscalibrated and overconfident on out-of-distribution (OOD) samples, MC-dropout provides well-calibrated confidence estimates. These improved uncertainty estimates are key to BiTTA's dual-path strategy: selecting the uncertain samples for feedback (BFA) and identifying confidently predicted ones for self-adaptation (ABA). We empirically validate the calibration benefits of MC-dropout in Section~\ref{sec:experiment}.

\vspace{-0.15cm}\paragraph{Dual-path optimization strategy.} With the proposed RL framework,  \method{} addresses the challenge of TTA with binary feedback (Figure~\ref{fig:overview}), utilizing (1) \textit{few samples with ground-truth binary feedback} and (2) \textit{many unlabeled samples with potentially noisy predictions} by two complementary strategies:
\begin{enumerate}
\vspace{-0.2cm}
    \item Binary Feedback-guided Adaptation on uncertain samples (BFA, Section~\ref{sec:binary-adaptation}): This strategy focuses on enhancing the model's areas of uncertainty. By selecting samples where the model is least confident and obtaining binary feedback on these, \method{} efficiently probes the boundaries of the model's current knowledge. 
    % This guided adaptation is crucial for adapting to distribution shifts, empowering the model to learn from its mistakes and improve its performance on challenging samples.

    \item Agreement-Based self-Adaptation on confident samples (ABA, Section~\ref{sec:unlabeled-adapt}): To complement the guided adaptation strategy, \method{} also leverages the model's existing knowledge through self-adaptation on confidently predicted samples. Without requiring additional feedback, ABA identifies confident samples by the agreement between the model's standard predictions and those obtained via MC-dropout. 
    % This self-adaptation mechanism helps to maintain the model's correct behaviors without requiring the oracle feedback.
\end{enumerate}

% \vspace{-0.3cm}
The synergy between BFA and ABA enables \method{} to effectively utilize both labeled and unlabeled samples. BFA drives exploration and adaptation to new patterns in the test distribution through binary feedback on uncertain samples. Concurrently, ABA maintains and refines existing knowledge through self-supervised adaptation on confident predictions. This dual-path optimization allows for effective adaptation across diverse challenging conditions.

\subsection{Binary Feedback-Guided Adaptation}\label{sec:binary-adaptation}

In \system{} settings where binary feedback is limited and costly, selecting samples to query and using them for effective model adaptation becomes crucial. To address this challenge, we propose Binary Feedback-guided Adaptation on uncertain samples (BFA). This approach refines the model's decision boundaries and improves its understanding of challenging data points through binary feedback guidance, enabling robust and efficient adaptation in test-time distribution shifts. 
% A key challenge in \system{} is identifying which samples are most informative for model updating, especially when only limited feedback can be obtained. To address this, \method{} implements Binary Feedback-guided Adaptation (BFA) on uncertain samples, focusing on areas where the model exhibits the highest uncertainty. 

\vspace{-0.15cm}\paragraph{Sample selection.}
To guide the adaptation, we focus on the uncertain samples, often the most informative for model improvement~\citep{settles2009active}. 
We quantify the sample-wise (un)certainty $C(x)$ as the MC-dropout softmax of the original model predicted class:
% approximates the stochastic prediction probability by applying dropout at test time. 
% Therefore, we define the sample-wise certainty $C(x)$ as:
\begin{equation}
    {C}(x) = \pi_\theta(y^*|x),
\end{equation}
where $y^* = \argmax_y f_\theta(y|x)$ is the deterministic class prediction and $\pi_\theta(y|x)$ is MC-dropout softmax confidence.

Then, we select the set of $k$ samples to get the binary feedback, noted as $\mathcal{S}_{\tt BFA}$. One straightforward strategy is to select the least confident samples:
\begin{equation}\label{eq:bfa}
    \mathcal{S}_{\tt BFA} = {\tt argsort}_x({C}(x))[:k].
\end{equation}
We further discuss the BFA sample selection strategy and its impact in Appendix~\ref{app:ablation}.

\vspace{-0.15cm}\paragraph{Reward function design.}
For these selected samples, we query the binary feedback $B(x, y)$ (correct/incorrect) and define the reward function $R_{\tt BFA}$ for feedback samples as:
\begin{equation}
    R_{\tt BFA}(x, y) = B(x, y) = 
    \begin{cases} 
        1 & \text{if $y$ is correct,} \\
        -1 & \text{otherwise.}
    \end{cases}
\end{equation}
This binary-feedback reward scheme provides a clear signal for model adaptation, encouraging the prediction probability of correct predictions and penalizing incorrect ones. By selectively applying this reward function to the most uncertain samples, BFA efficiently utilizes the limited labeling budget, maximizing the contribution of each queried label.

% By selectively querying feedback on these uncertain samples, BFA efficiently utilizes the limited labeling budget to focus on areas where the model's current knowledge is least reliable, generating a strong learning signal on uncertain predictions and facilitating effective adaptation to potential distribution shifts in the test data.

\begin{figure*}[ht]
    \begin{subfigure}[t]{0.48\linewidth }
    \centering
    \includegraphics[width=\linewidth ]{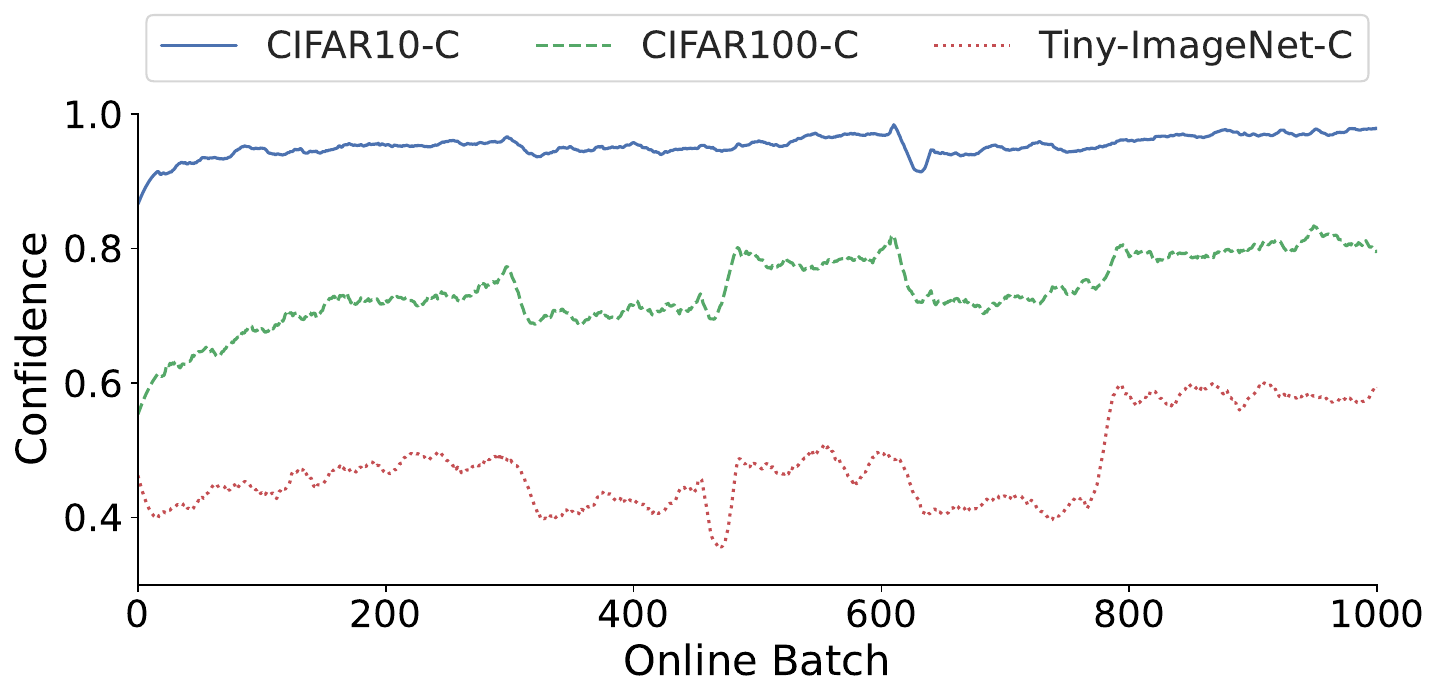}
    \caption{Average sample-wise confidence in three image corruption datasets.}\label{fig:conf_correct}
    \end{subfigure}
    \hspace{0.5cm}
    \begin{subfigure}[t]{0.48\linewidth }
    \centering
    \includegraphics[width=\linewidth ]{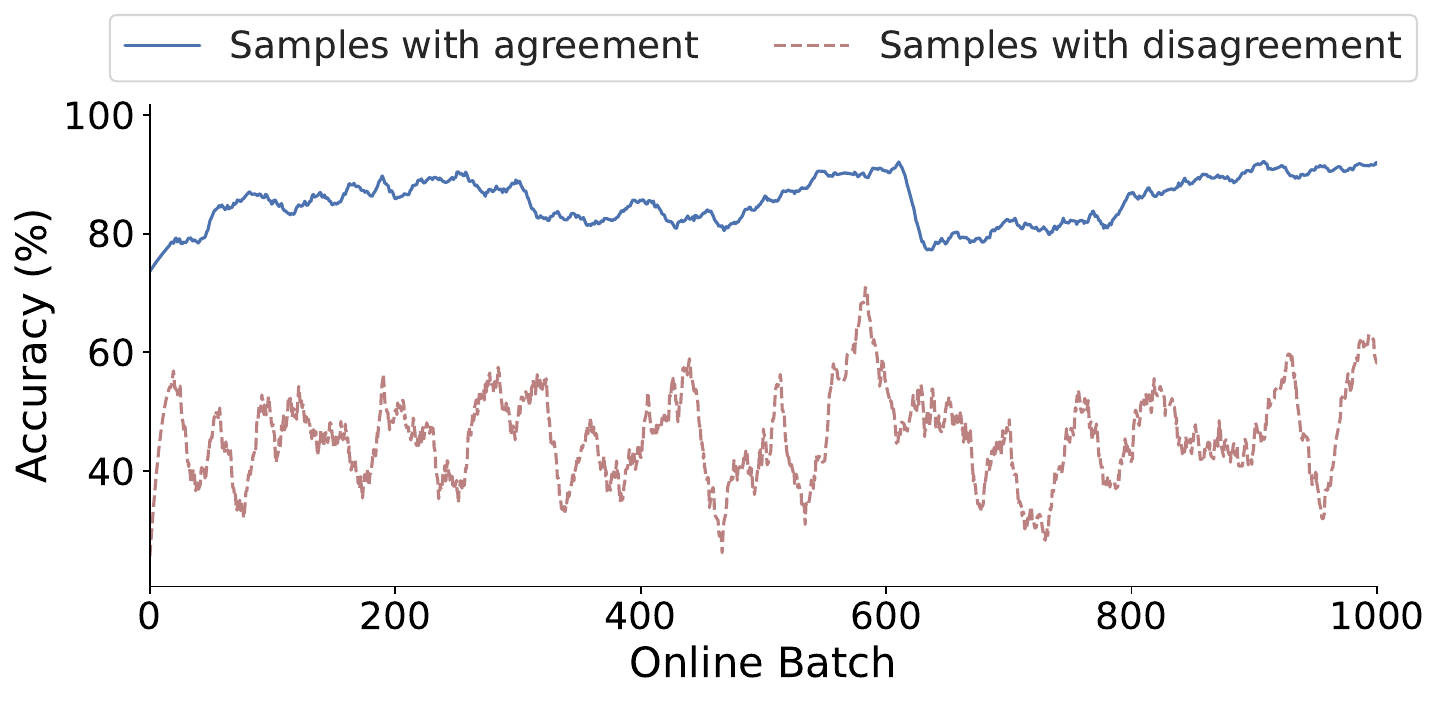}
    \caption{Average sample-wise accuracy of samples with agreement ($\mathcal{S}_{\tt ABA}$) and disagreement ($\mathcal{B} \setminus \mathcal{S}_{\tt ABA}$).}\label{fig:conf_agreement}
    % \vspace{-0.3cm}
    \end{subfigure}
    \vspace{-.1cm}
    \caption{Analysis of confidence and accuracy during online adaptation.  (a) Average sample-wise confidence over time and dataset, showing dynamic changes that challenge fixed thresholding methods. (b) Average sample-wise accuracy for samples with prediction agreement and disagreement on CIFAR10-C, demonstrating the effectiveness of agreement-based selection for confident samples.}
    \label{fig:conf}
    \vspace{-0.3cm}
    % \hfill
\end{figure*}

\subsection{Agreement-Based Self-Adaptation}\label{sec:unlabeled-adapt}

% Our second idea is complementing the binary feedback-guided adaptation by leveraging the model's confident predictions with many unlabeled samples.
% The key idea is \textbf{Agreement-Based self-Adaptation on confident samples (ABA)}. We calculate the agreement between the model and MC-dropout prediction and use these samples to reinforce the model's current knowledge without requiring additional oracle feedback. We define a set of confident samples with prediction  agreement $\mathcal{S}_{\tt ABA}$ as:

To complement the binary feedback-guided adaptation on uncertain samples, we propose leveraging the model's confident predictions on the remaining many unlabeled samples. This approach, which we call Agreement-Based self-Adaptation (ABA), aims to reinforce the model's current knowledge without requiring additional oracle feedback.

\vspace{-0.15cm}\paragraph{Sample selection.}
The key idea behind ABA is to identify samples where the model's standard prediction agrees with its MC-dropout prediction. We consider these samples ``confident'' and use them for self-adaptation. Formally, we define the set of confident samples $\mathcal{S}_{\tt ABA}$ as:
\begin{equation}\label{eq:agreement}
    \mathcal{S}_{\tt ABA} = \{x \in \mathcal{B} \setminus \mathcal{S}_{\tt BFA} \mid y^* = \argmax_y \pi_\theta(y|x)\},
\end{equation}
where $\mathcal{B}$ is the entire batch of test samples, $\mathcal{S}_{\tt BFA}$ is the set of samples selected for active feedback, $y^*$ is the original class prediction, and $\pi_\theta(y|x)$ is the MC-dropout softmax. 

Unlike existing TTA methods that rely on fixed confidence thresholds~\citep{eata, sar, sotta}, our approach can dynamically select confident samples based on the agreement between standard and MC-dropout predictions.
Figure~\ref{fig:conf_correct} illustrates the dynamic nature of prediction confidences during distribution shifts---highlighting the need for dynamic sample selection. To demonstrate the effectiveness of ABA further, we compare our agreement-based approach with thresholding strategies (Figure~\ref{fig:conf_threshold} in Appendix~\ref{app:ablation}). The results motivate our dynamic agreement-based selection over static confidence thresholding.

Furthermore, our method effectively identifies confident samples for self-adaptation. Figure~\ref{fig:conf_agreement} demonstrates the stable accuracies in samples with agreement, while samples with disagreement show unstable and low accuracies.
This originates from the prediction agreement of indicating robustness and reliability via the consistency in model outputs across different dropout masks. By leveraging this consistency, ABA can reliably select confident samples for effective self-adaptation.

% The key strength of ABA lies in its ability to automatically adjust to shifting confidence levels. This makes it particularly well-suited for scenarios with continuous distribution changes, a common challenge in real-world applications. 

\vspace{-0.15cm}\paragraph{Reward function design.}
We now incorporate these samples into our reinforcement learning framework. We introduce a self-feedback reward function $R_{\tt ABA}$ for unlabeled samples. This reward encourages the model to maintain its predictions on confident samples while discarding the adaptation on unreliable ones. Formally, we define $R_{\tt ABA}$ as:
\begin{equation}
    R_{\tt ABA}(x, y) = 
    \begin{cases} 
        1 & \text{if} \; x \in \mathcal{S}_{\tt ABA},\\
        0 & \text{otherwise.}
    \end{cases}
\end{equation}
By incorporating this adaptive prediction agreement strategy, ABA enhances the learning process by maintaining the knowledge of confident predictions.
While prediction disagreement suggests uncertainty, our analysis shows disagreement samples exhibit unstable accuracy rather than consistent errors (Figure~\ref{fig:conf_agreement}).
Therefore, ABA assigns zero rewards to them rather than penalizing (as in BFA), preventing potentially harmful adaptation from noisy signals. This conservative approach is especially valuable in TTA scenarios where distribution shift may be partial or gradual, where most of the model's existing knowledge remains relevant.
% ABA is especially valuable in TTA scenarios where the distribution shift may be partial or gradual, where most of the model's existing knowledge remains relevant.

\subsection{\method{} Algorithm}\label{sec:bitta}

Our proposed \method{} algorithm unifies binary feedback-guided adaptation (BFA, Section~\ref{sec:binary-adaptation}) and agreement-based self-adaptation (ABA, Section~\ref{sec:unlabeled-adapt}) into a dual-path optimization framework. This design enables robust and efficient test-time adaptation under distribution shifts while maintaining stability across time. Here, we detail the algorithmic formulation and its practical implementation using memory-based sample buffers.

\vspace{-0.15cm}\paragraph{Combined objective.}
We formulate the overall objective as a weighted sum of expected rewards from BFA and ABA:
\begin{equation}
    J(\theta) = \alpha \mathbb{E}_{x \in \mathcal{S}_{\text{BFA}}} \left[R_{\text{BFA}}(x, y)\right] + \beta \mathbb{E}_{x \in \mathcal{S}_{\text{ABA}}} \left[R_{\text{ABA}}(x, y)\right],
    \label{eq:total_reward}
\end{equation}
where $\alpha$ and $\beta$ balance BFA and ABA.
We set $\alpha = \beta = 1$ in all experiments and analyze the impact of these hyperparameters in Figure~\ref{fig:ablation_lambda} (Appendix~\ref{app:ablation}).

The policy gradient is estimated via \textsc{Reinforce}~\cite{williams1992simple}:
\begin{align}
    \nabla_\theta J(\theta) = 
    &\; \alpha \mathbb{E}_{x \in \mathcal{S}_{\text{BFA}}} \left[ R_{\text{BFA}}(x, y) \nabla_\theta \log \pi_\theta(y|x) \right] \nonumber \\
    &+ \beta \mathbb{E}_{x \in \mathcal{S}_{\text{ABA}}} \left[ \nabla_\theta \log \pi_\theta(y|x) \right],
    \label{eq:policy_gradient}
\end{align}
where $\pi_\theta(y|x)$ is approximated using MC-dropout.

\vspace{-0.15cm}\paragraph{Practical implementation.}
We implement the policy gradient in Equation~\ref{eq:policy_gradient} using two FIFO memory pools: $\mathcal{M}_{\texttt{C}}$ for correct predictions and $\mathcal{M}_{\texttt{I}}$ for incorrect ones. Each memory stores up to a batch-sized number of samples for computational stability and efficiency. Additionally, we use $\mathcal{S}_{\text{ABA}}$, confident unlabeled samples identified via agreement on the current batch. 

For each input $x$ with predicted label $y^*$, we optimize the MC-dropout softmax probability $\pi_\theta(y^*|x)$ by minimizing or maximizing the cross-entropy depending on its type. The overall loss is defined as:
\begin{align}
\mathcal{L}_{\text{BiTTA}} = &\; \alpha \cdot \frac{1}{|\mathcal{M}_{\texttt{C}}|} \sum_{x \in \mathcal{M}_{\texttt{C}}} \left(-\log \pi_\theta(y^*|x) \right) \nonumber \\
&+ \alpha \cdot \frac{1}{|\mathcal{M}_{\texttt{I}}|} \sum_{x \in \mathcal{M}_{\texttt{I}}} \left(+\log \pi_\theta(y^*|x) \right) \nonumber \\
&+ \beta \cdot \frac{1}{|\mathcal{S}_{\text{ABA}}|} \sum_{x \in \mathcal{S}_{\text{ABA}}} \left(-\log \pi_\theta(y^*|x) \right).
\label{eq:bitta}
\end{align}

\begin{algorithm}[t]
\caption{\method{} Algorithm }
\label{alg:tta}
\begin{algorithmic}[1]
\setstretch{1.1}
\STATE \textbf{Input:} Model $f_\theta$, MC-dropout $\pi_\theta$, batch $\mathcal{B}$, correct memory $\mathcal{M}_C$, incorrect memory $\mathcal{M}_I$, budget $k$, epochs $E$, learning rate $\eta$, balancing hyperparameters $\alpha, \beta$
\vspace{0.2em}
\STATE \texttt{\textcolor{gray}{\# Binary Feedback Sample Selection}}
\STATE $Y^* \gets \arg\max f_\theta(* \!\mid\! \mathcal{B})$ \hfill \textcolor{gray}{// Deterministic prediction}
\STATE $I_{\tt BFA} \gets \texttt{argsort}_x(\pi_\theta(Y^* \!\mid\! \mathcal{B}))[:k]$ \hfill \textcolor{gray}{// Equation~\ref{eq:bfa}}
\STATE $\mathcal{S}_{\tt BFA} \gets \{(x_i, y^*_i) \mid i \in I_{\tt BFA}\}$ \hfill \textcolor{gray}{// BFA samples}
\FOR{each $(x, y)$ in $\mathcal{S}_{\tt BFA}$}
    \STATE $B(x, y) \gets$ Binary feedback from oracle
    \IF{$B(x, y) = 1$}
        \STATE $\mathcal{M}_C.\texttt{update}(x, y)$ \hfill \textcolor{gray}{// Store correct sample}
    \ELSE
        \STATE $\mathcal{M}_I.\texttt{update}(x, y)$ \hfill \textcolor{gray}{// Store incorrect sample}
    \ENDIF
\ENDFOR
\vspace{0.2em}
\STATE \texttt{\textcolor{gray}{\# Test-Time Adaptation}}
\STATE Update BN stats with $\mathcal{B}$ and freeze them 
\FOR{$e = 1$ to $E$}
    \STATE \texttt{\textcolor{gray}{\# BFA (Section~\ref{sec:binary-adaptation})}}
    \STATE $X_C, Y_C \gets \mathcal{M}_C$ \hfill \textcolor{gray}{// Get correct samples}
    \STATE $X_I, Y_I \gets \mathcal{M}_I$ \hfill \textcolor{gray}{// Get incorrect samples}
    \STATE $\mathcal{L}_{\tt BFA} \gets \ell_{\texttt{CE}}(\pi_\theta(* \mid X_C), Y_C) - \ell_{\texttt{CE}}(\pi_\theta(* \mid X_I), Y_I)$ 
    
    \vspace{0.2em}
    \STATE \texttt{\textcolor{gray}{\# ABA (Section~\ref{sec:unlabeled-adapt})}}
    \STATE $X_U \gets \mathcal{B} \setminus \mathcal{S}_{\tt BFA}$ \hfill \textcolor{gray}{// Get unlabeled samples}
    \STATE $R_{\tt ABA} \!\gets\! \mathbb{1}[\arg\max f_\theta(* \!\mid\! X_U\!) \!=\! \arg\max \pi_\theta(* \!\mid\! X_U\!)]$ 
    \STATE $X_{\tt ABA} \gets \{x \in X_U \mid R_{\tt ABA}(x) = 1\}$
    , \\ $Y_{\tt ABA} \gets \arg\max f_\theta(* \mid X_{\tt ABA})$ \hfill \textcolor{gray}{// Get $\mathcal{S}_{\text{ABA}}$}
    \STATE $\mathcal{L}_{\tt ABA} \gets \ell_{\texttt{CE}}(\pi_\theta(* \mid X_{\tt ABA}), Y_{\tt ABA})$

    \vspace{0.2em}
    \STATE \texttt{\textcolor{gray}{\# Final Update (Section~\ref{sec:bitta})}}
    \STATE $\mathcal{L}_{\tt BiTTA} \gets \alpha \cdot \mathcal{L}_{\tt BFA} + \beta \cdot \mathcal{L}_{\tt ABA}$
    \STATE $\theta \gets \theta - \eta \nabla_\theta \mathcal{L}_{\tt BiTTA}$ 
\ENDFOR
\end{algorithmic}
\end{algorithm}

Based on the unified objective $\mathcal{L}_{\tt BiTTA}$, Algorithm~\ref{alg:tta} outlines the full adaptation procedure. The algorithm proceeds in two phases. In the first phase, we select the top-$k$ uncertain samples (or random samples in large-scale additional studies) from the current batch based on MC-dropout confidence scores over the predicted class. We then query binary feedback from an oracle and update the corresponding FIFO memory: $\mathcal{M}_{\texttt{C}}$ (correct) or $\mathcal{M}_{\texttt{I}}$ (incorrect). This separation ensures balanced and stable adaptation, particularly during early rounds when labeled feedback is scarce or skewed.

In the second phase, we perform $E$ epochs of test-time adaptation. We first freeze the batch normalization (BN) statistics using the current batch. We then compute two types of adaptation losses: (i) BFA over $\mathcal{M}_{\texttt{C}}$ and $\mathcal{M}_{\texttt{I}}$ by minimizing and maximizing cross-entropy, respectively, and (ii) ABA over unlabeled samples with agreement by minimizing cross-entropy. The final objective $\mathcal{L}_{\tt BiTTA}$ combines both components and is optimized via gradient descent. Notably, the policy gradient $\nabla_\theta \log \pi_\theta(y|x)$ is implemented implicitly via backpropagation through cross-entropy on MC-dropout outputs, enabling a simple and effective optimization.

% We formulate a combined objective function that balances the rewards from both uncertain samples (guided by binary feedback) and confident samples (identified through prediction agreement). Formally, we define our total objective function $J(\theta)$ as:
% \begin{equation}
%     J(\theta) = \alpha \mathbb{E}_{x \in \mathcal{S}_{\tt BFA}}[R_{\tt BFA}(x, y)] + \beta \mathbb{E}_{x \in \mathcal{B} \setminus \mathcal{S}_{\tt BFA}}[R_{\tt ABA}(x, y)],
% \end{equation}
% where $\alpha, \beta$ are hyperparameters to control the relative contributions of BFA and ABA.

% Following the \textsc{Reinforce} algorithm, the gradient of our total objective is given by:
% \begin{equation}\label{eq:total_objective}
% \begin{split}
%     \nabla_\theta J(\theta) = \alpha \mathbb{E}_{x \in \mathcal{S}_{\tt BFA}}[R_{\tt BFA}(x, y) \nabla_\theta \log \pi_\theta(y|x)] \\
%     + \beta \mathbb{E}_{x \in \mathcal{B} \setminus \mathcal{S}_{\tt BFA}}[R_{\tt ABA}(x, y) \nabla_\theta \log \pi_\theta(y|x)].
% \end{split}
% \end{equation}

% This gradient estimation guides our parameter updates via stochastic gradient ascent, refining the model's performance on the evolving test distribution. 
% %We utilize a single value of $\alpha=2$ and $\beta=1$ for all experiments. 
% Further implementation details, including the hyperparameters, are in Appendix~\ref{app:exp_detail}.

\section{Experiments}\label{sec:experiment}

\begin{table*}[t]
\centering
\caption{Accuracy (\%) comparisons with TTA and active TTA baselines with binary feedback in corruption datasets (severity level 5). Notation * indicates the modified algorithm to utilize binary-feedback samples. B: TTA with binary feedback. Results outperforming all other baselines are highlighted in \textbf{bold} fonts. Averaged over three random seeds. Comparison with non-active TTAs and full-class active TTA are in Table~\ref{tab:app_original} in Appendix~\ref{app:add_result}.}
\label{tab:tinyimagenetc}
\footnotesize

\begin{subtable}{\textwidth}
    
\resizebox{\linewidth}{!}{%
\begin{tabularx}{1.3\textwidth}{c|l   *{16}Y}
\toprule
& & \multicolumn{3}{c}{Noise} & \multicolumn{4}{c}{Blur} & \multicolumn{4}{c}{Weather} & \multicolumn{4}{c}{Digital} &  \\
& & \multicolumn{15}{l}{ %$t$ 
    \begin{tikzpicture}[overlay, remember picture]
      \draw[->, line width=0.2pt] (0,0.075) -- (19,0.075);
    \end{tikzpicture}
    } &  \\
\multirow{-3}{*}{Label} & {\multirow{-3}{*}{Method}} & Gau. & Shot & Imp. & Def. & Gla. & Mot. & Zoom & Snow & Fro. & Fog & Brit. & Cont. & Elas. & Pix. & JPEG & Avg. \\
\midrule
- & SrcValid & 25.97 & 33.19 & 24.71 & 56.73 & 52.02 & 67.37 & 64.80 & 77.97 & 67.01 & 74.14 & 91.51 & 33.90 & 76.62 & 46.38 & 73.23 & 57.23 \\
- & BN-Stats & 66.96 & 69.04 & 60.36 & {87.78} & 65.55 & 86.29 & 87.38 & 81.63 & 80.28 & 85.39 & 90.74 & 86.88 & 76.72 & 79.33 & 71.92 & 78.42 \\
B & TENT* & 75.25 & 80.71 & 73.21 & 85.42 & 68.89 & 82.95 & 85.48 & 81.96 & 82.99 & 83.18 & 88.88 & 86.78 & 75.37 & 80.77 & 75.59 & 80.49 \\
B & EATA* & 73.22 & 74.16 & 65.72 & 76.50 & 62.47 & 74.62 & 79.65 & 76.42 & 77.31 & 79.70 & 85.37 & 83.66 & 69.52 & 77.12 & 72.16 & 75.17 \\
B & SAR* & 71.57 & 78.62 & 73.33 & \textbf{88.98} & 73.30 & 87.98 & 89.72 & 86.00 & 87.09 & 87.86 & 92.46 & 90.00 & 82.07 & 86.69 & 80.96 & 83.78 \\
B & CoTTA* & 66.97 & 69.04 & 60.35 & 87.77 & 65.54 & 86.29 & 87.38 & 81.63 & 80.28 & 85.40 & 90.73 & 86.87 & 76.74 & 79.35 & 71.92 & 78.42 \\
B & RoTTA* & 64.49 & 69.69 & 63.89 & 86.29 & 69.48 & 86.78 & 89.23 & 85.21 & 85.39 & 87.48 & 92.02 & 87.09 & 81.76 & 85.66 & 80.18 & 80.98 \\
B & SoTTA* & 71.39 & 79.27 & 70.58 & 85.07 & 68.39 & 84.03 & 87.27 & 83.47 & 84.90 & 85.55 & 90.81 & 86.18 & 78.26 & 83.41 & 76.94 & 81.03 \\
B & SimATTA* & 70.21 & 81.67 & 71.49 & 79.59 & 69.41 & 82.15 & 87.28 & 83.90 & 86.89 & 86.49 & 91.51 & 83.40 & 77.94 & 83.81 & 82.25 & 81.09 \\
B & \cellcolor[HTML]{DCFFDC}\method{} & \cellcolor[HTML]{DCFFDC}\textbf{76.78} & \cellcolor[HTML]{DCFFDC}\textbf{84.24} & \cellcolor[HTML]{DCFFDC}\textbf{78.75} & \cellcolor[HTML]{DCFFDC}87.51 & \cellcolor[HTML]{DCFFDC}\textbf{77.39} & \cellcolor[HTML]{DCFFDC}\textbf{88.38} & \cellcolor[HTML]{DCFFDC}\textbf{91.36} & \cellcolor[HTML]{DCFFDC}\textbf{89.42} & \cellcolor[HTML]{DCFFDC}\textbf{90.72} & \cellcolor[HTML]{DCFFDC}\textbf{90.30} & \cellcolor[HTML]{DCFFDC}\textbf{94.65} & \cellcolor[HTML]{DCFFDC}\textbf{92.62} & \cellcolor[HTML]{DCFFDC}\textbf{86.15} & \cellcolor[HTML]{DCFFDC}\textbf{92.42} & \cellcolor[HTML]{DCFFDC}\textbf{87.24} & \cellcolor[HTML]{DCFFDC}\textbf{87.20} \\
\bottomrule
\end{tabularx}
}
\vspace{0.05cm}
\caption{CIFAR10-C.}
\end{subtable}

\begin{subtable}{\textwidth}
    
\resizebox{\linewidth}{!}{%
\begin{tabularx}{1.3\textwidth}{c|l  *{16}Y}
\toprule
& & \multicolumn{3}{c}{Noise} & \multicolumn{4}{c}{Blur} & \multicolumn{4}{c}{Weather} & \multicolumn{4}{c}{Digital} &  \\
& & \multicolumn{15}{l}{ %$t$ 
    \begin{tikzpicture}[overlay, remember picture]
      \draw[->, line width=0.2pt] (0,0.075) -- (19,0.075);
    \end{tikzpicture}
    } &  \\
{\multirow{-3}{*}{Label}} & {\multirow{-3}{*}{Method}} & Gau. & Shot & Imp. & Def. & Gla. & Mot. & Zoom & Snow & Fro. & Fog & Brit. & Cont. & Elas. & Pix. & JPEG & Avg. \\
\midrule
- & SrcValid & 10.63 & 12.14 & \hphantom{0}7.17 & 34.86 & 19.58 & 44.09 & 41.94 & 46.34 & 34.22 & 41.08 & 67.31 & 18.47 & 50.36 & 24.91 & 44.56 & 33.18 \\
- & BN-Stats & 39.23 & 40.75 & 34.10 & 66.14 & 42.46 & 63.57 & 64.82 & 53.81 & 53.49 & 58.15 & 68.22 & 64.48 & 53.88 & 56.63 & 45.17 & 53.66 \\
B & TENT* & 49.70 & 51.52 & 39.35 & 44.63 & 28.42 & 27.26 & 24.74 & 14.74 & 10.19 & \hphantom{0}6.44 & \hphantom{0}4.89 & \hphantom{0}3.01 & \hphantom{0}2.99 & \hphantom{0}2.87 & \hphantom{0}2.62 & 20.89 \\
B & EATA* & 10.80 & \hphantom{0}2.75 & \hphantom{0}2.46 & \hphantom{0}1.88 & \hphantom{0}1.68 & \hphantom{0}1.80 & \hphantom{0}1.67 & \hphantom{0}1.56 & \hphantom{0}1.37 & \hphantom{0}1.22 & \hphantom{0}1.30 & \hphantom{0}1.06 & \hphantom{0}1.39 & \hphantom{0}1.42 & \hphantom{0}1.23 & \hphantom{0}2.24 \\
B & SAR* & 46.57 & 55.41 & 48.54 & \textbf{66.29} & 51.01 & \textbf{65.27} & \textbf{68.49} & 60.72 & 62.35 & 63.35 & 71.08 & \textbf{69.50} & 59.82 & 65.55 & 56.48 & 60.70 \\
B & CoTTA* & 39.22 & 40.76 & 34.10 & 66.11 & 42.48 & 63.57 & 64.83 & 53.80 & 53.48 & 58.15 & 68.22 & 64.48 & 53.85 & 56.66 & 45.17 & 53.66 \\
B & RoTTA* & 36.68 & 40.73 & 36.00 & 62.54 & 44.58 & 62.90 & 66.93 & 58.09 & 59.71 & 60.23 & 70.33 & 62.67 & 59.12 & 63.55 & 53.83 & 55.86 \\
B & SoTTA* & 44.39 & 38.67 & 23.70 & 29.92 & 20.45 & 26.89 & 31.05 & 24.68 & 26.63 & 25.25 & 33.69 & 16.99 & 24.57 & 27.67 & 24.82 & 27.96 \\
B & SimATTA* & 28.93 & 41.54 & 28.94 & 39.79 & 34.83 & 49.11 & 55.42 & 46.59 & 51.06 & 48.83 & 60.03 & 34.67 & 44.27 & 45.19 & 46.22 & 43.69 \\
B & \cellcolor[HTML]{DCFFDC}\method{} & \cellcolor[HTML]{DCFFDC}\textbf{50.12} & \cellcolor[HTML]{DCFFDC}\textbf{58.34} & \cellcolor[HTML]{DCFFDC}\textbf{52.07} & \cellcolor[HTML]{DCFFDC}63.27 & \cellcolor[HTML]{DCFFDC}\textbf{52.70} & \cellcolor[HTML]{DCFFDC}63.80 & \cellcolor[HTML]{DCFFDC}68.16 & \cellcolor[HTML]{DCFFDC}\textbf{62.65} & \cellcolor[HTML]{DCFFDC}\textbf{65.39} & \cellcolor[HTML]{DCFFDC}\textbf{63.79} & \cellcolor[HTML]{DCFFDC}\textbf{71.26} & \cellcolor[HTML]{DCFFDC}68.97 & \cellcolor[HTML]{DCFFDC}\textbf{63.93} & \cellcolor[HTML]{DCFFDC}\textbf{69.45} & \cellcolor[HTML]{DCFFDC}\textbf{63.38} & \cellcolor[HTML]{DCFFDC}\textbf{62.49} \\
\bottomrule
\end{tabularx}
}
\vspace{0.05cm}
\caption{CIFAR100-C.}
\end{subtable}

\begin{subtable}{\textwidth}
    
\resizebox{\linewidth}{!}{%
\begin{tabularx}{1.3\textwidth}{c|l  *{16}Y}
\toprule
& & \multicolumn{3}{c}{Noise} & \multicolumn{4}{c}{Blur} & \multicolumn{4}{c}{Weather} & \multicolumn{4}{c}{Digital} &  \\
& & \multicolumn{15}{l}{ %$t$ 
    \begin{tikzpicture}[overlay, remember picture]
      \draw[->, line width=0.2pt] (0,0.075) -- (19,0.075);
    \end{tikzpicture}
    } &  \\
{\multirow{-3}{*}{Label}} & {\multirow{-3}{*}{Method}} & Gau. & Shot & Imp. & Def. & Gla. & Mot. & Zoom & Snow & Fro. & Fog & Brit. & Cont. & Elas. & Pix. & JPEG & Avg. \\
\midrule
- & SrcValid & \hphantom{0}6.66 & \hphantom{0}8.91 & \hphantom{0}3.68 & 16.03 & 10.05 & 26.62 & 27.16 & 29.29 & 33.40 & 11.26 & 30.76 & \hphantom{0}1.96 & 27.83 & 40.67 & 47.96 & 21.48 \\
- & BN-Stats & 32.28 & 34.27 & 22.21 & 32.63 & 22.02 & 44.56 & 46.11 & 39.31 & 43.27 & 31.66 & 46.01 & 10.19 & 43.51 & 52.14 & 50.62 & 36.72 \\
B & TENT* & \textbf{36.80} & 35.24 & 17.71 & \hphantom{0}8.21 & \hphantom{0}3.33 & \hphantom{0}3.30 & \hphantom{0}2.91 & \hphantom{0}2.36 & \hphantom{0}2.15 & \hphantom{0}1.99 & \hphantom{0}1.99 & \hphantom{0}1.43 & \hphantom{0}1.88 & \hphantom{0}1.94 & \hphantom{0}1.88 & \hphantom{0}8.21 \\
B & EATA* & 35.19 & 37.98 & 25.35 & \textbf{35.82} & 24.84 & 47.04 & 48.19 & 42.10 & 45.53 & 36.67 & 48.79 & \hphantom{0}9.16 & 44.71 & 53.43 & 51.60 & 39.09 \\
B & SAR* & 33.90 & 38.56 & 27.84 & {35.49} & \textbf{26.70} & 47.14 & 48.41 & 41.54 & 45.49 & 37.25 & 49.53 & \textbf{14.85} & 46.63 & 52.49 & 51.41 & 39.82 \\
B & CoTTA* & 32.27 & 34.26 & 22.21 & 32.62 & 22.04 & 44.56 & 46.11 & 39.30 & 43.27 & 31.67 & 46.01 & 10.18 & 43.51 & 52.18 & 50.60 & 36.72 \\
B & RoTTA* & 31.54 & 35.42 & 23.98 & 32.80 & 24.02 & 45.16 & 47.01 & 40.70 & 44.57 & 33.07 & 47.57 & 14.59 & 44.65 & 49.89 & 49.49 & 37.63 \\
B & SoTTA* & 34.24 & 33.48 & 18.21 & 17.96 & 10.63 & 15.41 & 12.75 & \hphantom{0}9.46 & \hphantom{0}7.35 & \hphantom{0}5.97 & \hphantom{0}7.70 & \hphantom{0}1.64 & \hphantom{0}5.96 & \hphantom{0}6.13 & \hphantom{0}5.70 & 12.84 \\
B & SimATTA* & 16.76 & 24.53 & 14.51 & 17.92 & 14.72 & 30.63 & 34.98 & 25.21 & 34.89 & 17.91 & 35.13 & \hphantom{0}1.68 & 33.33 & 42.04 & 46.72 & 26.06 \\
B & \cellcolor[HTML]{DCFFDC}\method{} & \cellcolor[HTML]{DCFFDC}34.84 & \cellcolor[HTML]{DCFFDC}\textbf{39.88} & \cellcolor[HTML]{DCFFDC}\textbf{28.56} & \cellcolor[HTML]{DCFFDC}35.37 & \cellcolor[HTML]{DCFFDC}26.65 & \cellcolor[HTML]{DCFFDC}\textbf{48.41} & \cellcolor[HTML]{DCFFDC}\textbf{49.57} & \cellcolor[HTML]{DCFFDC}\textbf{43.62} & \cellcolor[HTML]{DCFFDC}\textbf{47.90} & \cellcolor[HTML]{DCFFDC}\textbf{39.53} & \cellcolor[HTML]{DCFFDC}\textbf{50.95} & \cellcolor[HTML]{DCFFDC}12.27 & \cellcolor[HTML]{DCFFDC}\textbf{47.18} & \cellcolor[HTML]{DCFFDC}\textbf{54.01} & \cellcolor[HTML]{DCFFDC}\textbf{54.06} & \cellcolor[HTML]{DCFFDC}\textbf{40.85} \\
\bottomrule
\end{tabularx}
}
\vspace{0.05cm}
\caption{Tiny-ImageNet-C.}
\end{subtable}

\vspace{-0.15cm}
\end{table*}

We present our experimental setup and results across various scenarios. % in \system{} setting. 
% We evaluate the performance of \method{} against state-of-the-art baselines, ensuring fairness by providing an equal amount of binary-feedback samples.
Additional experiments, results, and details are provided in Appendices~\ref{app:ablation}, \ref{app:add_result}, and \ref{app:exp_detail}.

\vspace{-0.15cm}\paragraph{Baselines.} We evaluated \method{} against a comprehensive set of baselines, including source validation (SrcValid) and seven state-of-the-art TTA methods: BN-Stats~\citep{bnstats}, TENT~\citep{tent}, EATA~\citep{eata}, SAR~\citep{sar}, CoTTA~\citep{cotta}, RoTTA~\citep{rotta}, and SoTTA~\citep{sotta}. To ensure a fair comparison, we incorporate an equal number of random binary-feedback data into TTA baselines by adding correct-sample loss (cross-entropy) and incorrect-sample loss (complementary label loss~\citet{kim2019nlnl}). We also included SimATTA~\citep{atta} as an active TTA baseline, adapting it to use binary-feedback data by incorporating a complementary loss for negative samples. The non-active and full-class active TTA accuracies are reported in Appendix~\ref{app:add_result}. 

%, ELPT~\citep{elpt}, and MHPL~\citep{mhpl} as active domain adaptation baselines.

\vspace{-0.15cm}\paragraph{Dataset.} To evaluate the robustness of \method{} across various domain shifts, we used standard image corruption datasets CIFAR10-C, CIFAR100-C, and Tiny-ImageNet-C~\citep{cifarc}. Additionally, we conducted experiments on the PACS dataset~\citep{pacs}, which is commonly used for domain adaptation tasks.
% For most experiments, we pre-trained the source model on the source domain while adapting and evaluating the model on the test-time domains. 
To closely simulate real-world scenarios with evolving distribution shifts, we implemented a continual TTA~\citep{cotta} where corruption continuously changes.

\vspace{-0.15cm}\paragraph{Settings and hyperparameters.} We configured \method{} to operate with minimal labeling effort, using only three binary feedback samples within each 64-sample test batch, accounting for less than 5\%. We utilize a single value of balancing hyperparameters $\alpha = 2$ and $\beta = 1$ for \method{} in all experiments.
A comprehensive list of settings and hyperparameters is provided in Appendix~\ref{app:exp_detail}.

\begin{table*}[t]
\centering
\caption{Accuracy (\%) comparisons with TTA and active TTA baselines with binary feedback in PACS. The domain-wise data stream is a continual TTA setting, and the mixed data stream shuffled all domains randomly, where we report the cumulative accuracy at four adaptation points. Notation * indicates the modified algorithm to utilize binary-feedback samples. B: TTA with binary feedback. Results outperforming all other baselines are highlighted in \textbf{bold} fonts. Comparison with non-active TTAs and full-class active TTA are in Table~\ref{tab:app_original_pacs} in Appendix~\ref{app:add_result}.}

\label{tab:pacs_vlcs}
\scriptsize	
\begin{tabularx}{\textwidth}{c|l *{8}Y}

\toprule

& & \multicolumn{4}{c}{Domain-wise data stream} & \multicolumn{4}{c}{Mixed data stream} \\ % \cmidrule(lr){7-7}\cmidrule(lr){12-12}
& & \multicolumn{4}{l}{ %$t$ 
    \begin{tikzpicture}[overlay, remember picture]
      \draw[->, line width=0.2pt] (0,0.075) -- (7,0.075);
    \end{tikzpicture}
    } & 
    \multicolumn{4}{l}{ %$t$ 
    \begin{tikzpicture}[overlay, remember picture]
      \draw[->, line width=0.2pt] (0,0.075) -- (7,0.075);
    \end{tikzpicture}
    } \\
{\multirow{-3}{*}{Label}} & {\multirow{-3}{*}{Method}} & Art & Cartoon & Sketch & Avg & 25\% & 50\% & 75\% & 100\%(Avg) \\
\midrule
- & Src & 59.38 \scalebox{\std}{±0.00} & 27.94 \scalebox{\std}{±0.21} & 42.96 \scalebox{\std}{±0.01} & 43.43 \scalebox{\std}{±0.07} & 42.74 \scalebox{\std}{±1.13} & 42.80 \scalebox{\std}{±0.22} & 42.64 \scalebox{\std}{±0.30} & 42.77 \scalebox{\std}{±0.01} \\
- & BN Stats & 67.87 \scalebox{\std}{±0.18} & 63.48 \scalebox{\std}{±0.88} & 54.07 \scalebox{\std}{±0.36} & 61.81 \scalebox{\std}{±0.18} & 59.09 \scalebox{\std}{±0.29} & 58.28 \scalebox{\std}{±0.08} & 58.05 \scalebox{\std}{±0.22} & 57.82 \scalebox{\std}{±0.20} \\
B & TENT* & 71.26 \scalebox{\std}{±0.44} & 67.71 \scalebox{\std}{±0.89} & 51.57 \scalebox{\std}{±1.73} & 63.51 \scalebox{\std}{±0.33} & \textbf{60.72 \scalebox{\std}{±0.71}} & 58.86 \scalebox{\std}{±0.53} & 57.54 \scalebox{\std}{±0.33} & 56.21 \scalebox{\std}{±0.15} \\
B & EATA* & 68.67 \scalebox{\std}{±0.38} & 65.31 \scalebox{\std}{±0.78} & 59.05 \scalebox{\std}{±0.27} & 64.34 \scalebox{\std}{±0.22} & 59.58 \scalebox{\std}{±0.10} & 59.15 \scalebox{\std}{±0.55} & 59.26 \scalebox{\std}{±0.36} & 59.39 \scalebox{\std}{±0.06} \\
B & SAR* & 67.95 \scalebox{\std}{±0.20} & 63.66 \scalebox{\std}{±0.81} & 55.35 \scalebox{\std}{±0.39} & 62.32 \scalebox{\std}{±0.12} & 59.17 \scalebox{\std}{±0.14} & 58.57 \scalebox{\std}{±0.21} & 58.42 \scalebox{\std}{±0.06} & 58.41 \scalebox{\std}{±0.08} \\
B & CoTTA* & 67.87 \scalebox{\std}{±0.18} & 63.47 \scalebox{\std}{±0.90} & 54.07 \scalebox{\std}{±0.36} & 61.80 \scalebox{\std}{±0.19} & 59.10 \scalebox{\std}{±0.32} & 58.29 \scalebox{\std}{±0.09} & 58.06 \scalebox{\std}{±0.23} & 57.83 \scalebox{\std}{±0.22} \\
B & RoTTA* & 64.73 \scalebox{\std}{±0.20} & 55.14 \scalebox{\std}{±1.91} & 56.05 \scalebox{\std}{±0.72} & 58.64 \scalebox{\std}{±0.50} & 55.50 \scalebox{\std}{±1.30} & 52.68 \scalebox{\std}{±0.64} & 51.45 \scalebox{\std}{±0.56} & 50.10 \scalebox{\std}{±0.33} \\
B & SoTTA* & 69.73 \scalebox{\std}{±0.43} & 42.48 \scalebox{\std}{±2.31} & 46.07 \scalebox{\std}{±2.00} & 52.76 \scalebox{\std}{±0.84} & 54.33 \scalebox{\std}{±3.59} & 52.89 \scalebox{\std}{±3.95} & 53.09 \scalebox{\std}{±3.78} & 53.37 \scalebox{\std}{±2.81} \\
B & SimATTA* & \scalebox{0.9}{55.83 \scalebox{0.7}{±16.69}} & 59.68 \scalebox{\std}{±7.98} & 72.40 \scalebox{\std}{±4.51} & 62.63 \scalebox{\std}{±9.63} & 59.34 \scalebox{\std}{±2.78} & 63.81 \scalebox{\std}{±0.68} & 67.09 \scalebox{\std}{±0.34} & 69.21 \scalebox{\std}{±0.11} \\
B & \cellcolor[HTML]{DCFFDC}\method{} & \cellcolor[HTML]{DCFFDC}\textbf{73.86 \scalebox{\std}{±3.76}} & \cellcolor[HTML]{DCFFDC}\textbf{76.81 \scalebox{\std}{±2.45}} & \cellcolor[HTML]{DCFFDC}\textbf{76.03 \scalebox{\std}{±1.61}} & \cellcolor[HTML]{DCFFDC}\textbf{75.57 \scalebox{\std}{±0.93}} & \cellcolor[HTML]{DCFFDC}59.65 \scalebox{\std}{±0.70} & \cellcolor[HTML]{DCFFDC}\textbf{64.70 \scalebox{\std}{±0.78}} & \cellcolor[HTML]{DCFFDC}\textbf{69.23 \scalebox{\std}{±0.17}} & \cellcolor[HTML]{DCFFDC}\textbf{72.18 \scalebox{\std}{±0.38}} \\
\bottomrule

\end{tabularx}
\end{table*}

\begin{figure*}[t]
    \centering
    \begin{minipage}[t]{0.34\textwidth}

        \centering
        \includegraphics[width=\linewidth]{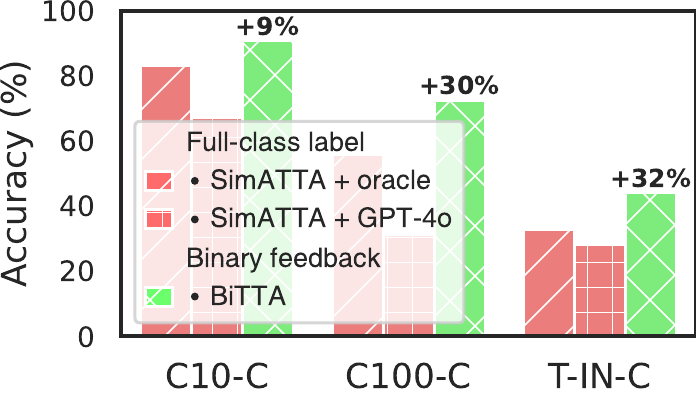}
        \caption{Accuracy (\%) with full-class feedback (SimATTA) and binary-feedback (\method{}) and under the equal total labeling cost. GPT-4o is used as a foundational model to provide a full-class label. }
        \label{fig:acc_atta}

    \end{minipage}
    \hfill
    \begin{minipage}[t]{0.3\textwidth}

        \centering
        \includegraphics[width=\linewidth]{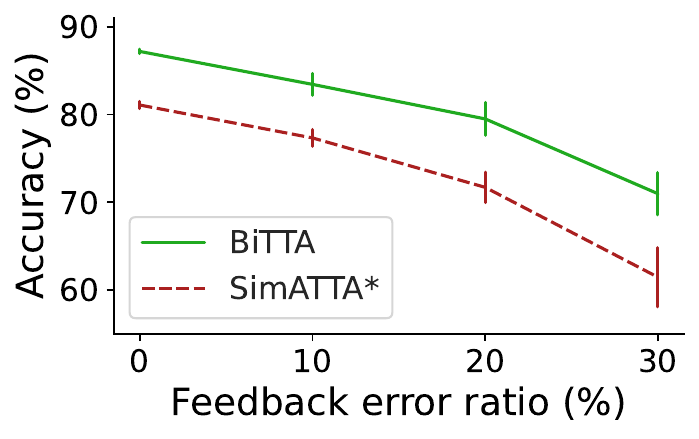}
        \caption{Accuracy (\%) varying the feedback error in CIFAR10-C. Binary feedback is flipped (\textit{correct $\leftrightarrow$ incorrect}) when a feedback error occurs. }
        \label{fig:feedback_error}

    \end{minipage}
    \hfill
    \begin{minipage}[t]{0.3\textwidth}
    % \captionsetup{labelfont={color=blue}}
        \centering
        \includegraphics[width=\linewidth]{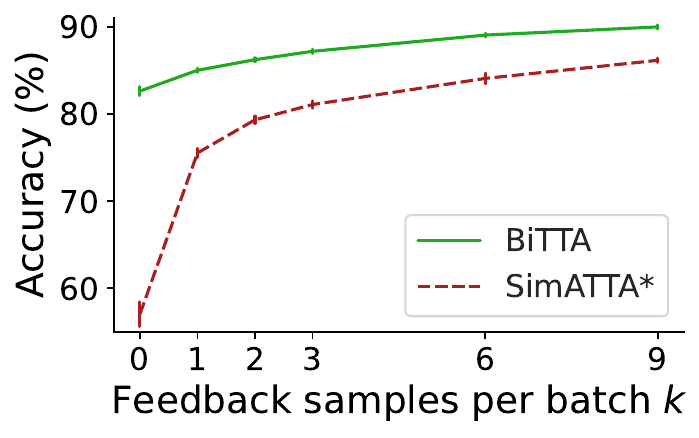}
        \caption{Accuracy (\%) varying the number of binary-feedback samples per batch ($k$) in CIFAR10-C. Zero feedback sample is equivalent to conventional TTA.}\label{fig:ablation_acc_n}
    
    \end{minipage}
\end{figure*}

\vspace{-0.15cm}\paragraph{Overall result.}
As shown in Table~\ref{tab:tinyimagenetc}, \method{} consistently outperformed all TTA and active TTA baselines, showcasing its effectiveness in the proposed \textbf{\system{}} setting. Existing TTA methods, even when adapted to binary feedback, showed suboptimal results, as their fixed filtering strategies (e.g., EATA, SAR, SoTTA) struggle to cope with dynamic uncertainties under continuous distribution shifts. SimATTA, an active TTA baseline, also underperformed due to hard-thresholding and clustering with incorrect samples.

% \vspace{-0.15cm}\paragraph{Comparison with full-label active TTA.} 
To fairly evaluate binary feedback under labeling constraints, we further compared \method{} against full-class active TTA in two scenarios: (1) \textit{equal labeling cost} --- adjusted for information gain (Figure~\ref{fig:acc_atta}), and (2) \textit{equal number of full-class/binary-feedback samples} (Table~\ref{tab:app_original}). Scenarios are further explained in Appendix~\ref{app:compare_active}. In both scenarios, \method{} (with binary feedback) consistently outperformed SimATTA (with full-class labels), demonstrating that \textbf{binary feedback alone can drive more effective adaptation} than full-class supervision under limited interactions. With equal labeling cost, \method{} achieved up to \textbf{32\% higher accuracy}, highlighting its \textbf{superior cost-efficiency and robustness}.
% As shown in Figure~\ref{fig:acc_atta}, this advantage becomes even more pronounced in datasets with a large number of classes, where full-labeling is costly and error-prone. 

In Figure~\ref{fig:acc_atta}, replacing an oracle feedback with a foundational model (GPT-4o) significantly degraded SimATTA’s performance due to GPT-4o’s high classification error (40\% on average). This result underlines the current limitations of automatic feedback: while easily accessible, it is unreliable under distribution shifts. In contrast, our approach leverages \textbf{lightweight yet trustworthy binary feedback from an oracle}, offering a robust and cost-efficient alternative for real-world test-time adaptation.

We also evaluated \method{} on broader distribution shifts in domain-wise (continual TTA~\citep{cotta}) and mixed-stream scenarios~\citep{atta} for a domain generalization task. Table~\ref{tab:pacs_vlcs} shows that \method{} achieved the highest average accuracy across all baselines, demonstrating generalization beyond corruptions.

\noindent Overall, these results show that \textbf{\system{} is a practical and scalable TTA paradigm}, and \textbf{\method{} delivers state-of-the-art performance} under minimal supervision, outperforming full-class approaches in both accuracy and efficiency.
Additional results, including large-scale and synthetic datasets (e.g., ImageNet-C~\citep{cifarc}, ImageNet-R~\citep{imagenetr}, ColoredMNIST~\citep{arjovsky2019invariant}, VisDA-2021~\citep{visda2021}, and DomainNet~\citep{domainnet}) and challenging settings (e.g., non-iid and single-sample), are available in Appendix~\ref{app:add_result}.

\vspace{-0.15cm}\paragraph{Impact of feedback errors.}
We assumed the binary feedback provided by the oracle contained no labeling errors. In practice, user feedback might include labeling errors by shifting the binary feedback between \textit{correct} and \textit{incorrect}. We examine the impact of binary-feedback error compared with the active TTA baseline. Figure~\ref{fig:feedback_error} shows that SimATTA has consistently low accuracy under labeling errors by relying on the noisy labeled samples without utilizing the unlabeled samples. In contrast, \method{} combines many confident unlabeled samples with labeled samples to reduce the impact of labeling errors, thus outperforming SimATTA.

\vspace{-0.15cm}\paragraph{Impact of number of feedback samples.}
We evaluated how the number of binary-feedback samples per batch ($k$) influences adaptation performance. As illustrated in Figures~\ref{fig:ablation_acc_n} and \ref{fig:acc_skip}, \method{} maintains high accuracy even with sparse active samples. The performance improves as $k$ increases, showcasing effective utilization of additional binary feedback. SimATTA shows a similar trend of increasing accuracy with more active samples, but the overall performance is consistently lower than \method{}.
This suggests that \method{} can effectively leverage additional feedback while maintaining stability in low budgets, indicating its potential for deployment in scenarios with varying labeling budgets.

\vspace{-0.15cm}\paragraph{Synergistic effect of adaptation strategies.} We compared \method{} against its components: Binary Feedback-guided Adaptation (BFA) and Agreement-based self-Adaptation (ABA). In CIFAR10-C, BFA-only adaptation achieved 58.90\% and ABA-only adaptation achieved 82.64\%, whereas \method{} achieved on average 87.20\% accuracy, consistently outperforming entire continual corruptions. The superior performance of the combined approach (\method{}) indicates that BFA and ABA complement each other to achieve robust accuracy.

\vspace{-0.15cm}\paragraph{Uncertainty calibration of MC-dropout.}
Beyond accuracy, we also analyze the calibration quality of MC-dropout compared with deterministic softmax outputs. Reliable calibration is essential for effective sample selection in both BFA and ABA. We compute the expected calibration error (ECE) across CIFAR10-C corruptions and find that MC-dropout achieves a substantially lower ECE of {0.062}, compared with {0.100} from softmax-based confidence, a {38\% reduction} in miscalibration. We further compare with various uncertainty estimation methods (e.g., augmentation, ensemble, and deterministic softmax confidence) in Table~\ref{tab:app_augment} (Appendix~\ref{app:ablation}), showcasing the importance of MC-dropout for the policy gradient modeling.

% \begin{wrapfigure}{r}{0.5\textwidth}
%     \centering
%     \vspace{-0.5cm}
%     \begin{subfigure}[t]{\linewidth}
%         \centering
%         \includegraphics[width=\linewidth]{fig/sample_selection.pdf}
%     \end{subfigure}
%     \caption{Accuracy (\%) by sample selection in PACS. Averaged over three random seeds.}
%     \label{fig:sample-selection}
%     \vspace{-0.4cm}
% \end{wrapfigure}

% \vspace{-0.15cm}\paragraph{Impact of active sample selection of \method{}.}
% We also investigated the impact of active selection strategies on the performance of \method{}.
% Figure~\ref{fig:sample-selection} presents the results of various sample selection methods, including our confidence-difference-based sample selection, random selection, maximum entropy~\citep{saito2020universal}, minimum confidence~\citep{sohn2020fixmatch}, and minimum energy~\citep{ood_energy}.
% The results show that random selection performed better than other uncertainty-based strategies, which aligns with findings in low-budget regimes in active learning~\citep{Zhu2018AddressingTI, Simoni2019RethinkingDA, Chandra2020OnIP, Hacohen2022ActiveLO}, where random selection outperforms when the initial labeled set is small or absent.
% However, our confidence-difference-based strategy outperformed all other methods, demonstrating the importance of selecting informative samples with significant differences between model and MC-dropout confidence. 

\section{Related Work}\label{sec:related_work}

\paragraph{Test-time adaptation.}

Test-time adaptation (TTA) improves model accuracy on distribution shift on the pre-trained model with only unlabeled test samples~\citep{tent}. Existing TTA focused on robust adaptation~\citep{ sar, note, rotta, cotta, lame, eata, sotta, park2024medbn} across various types of distribution shifts~\citep{sar, note, cotta, sotta, rdumb}. However, existing TTA methods suffer from adaptation failures during lifelong adaptation~\citep{rdumb}, highlighting the need for a few-sample guide for robust adaptation. Active test-time adaptation (ATTA)~\cite{atta} introduced a foundational analysis of active TTA setting, with a supervised learning scheme (SimATTA) using low-entropy source-like sample pseudo-labeling and active labeling from an incremental clustering algorithm. However, SimATTA is sensitive to the pre-trained model and selected active samples, as it does not leverage most unlabeled samples and only utilizes a few labeled samples. In contrast, \system{} utilizes a large set of unlabeled samples while guiding adaptation with binary-feedback samples, performing stable adaptation.

% \vspace{-0.15cm}\paragraph{Interactive machine learning.}

\vspace{-0.15cm}\paragraph{Reinforcement learning for model tuning.} Reinforcement learning (RL) has been successfully applied in various domains to incorporate non-differentiable rewards in the optimization process~\citep{zoph2017neural, yoon2020data, ouyang2022training, fan2024reinforcement, black2024training}.
For example, \citet{zoph2017neural} and \citet{yoon2020data} employ the \textsc{Reinforce} algorithm to use the accuracy of the validation dataset as a (non-differentiable) reward in neural architecture search or data valuation.
In the domain of natural language processing, reinforcement learning with human feedback (RLHF)~\cite{ouyang2022training} has gained prominence for fine-tuning large language models,
% Such a recipe has been extended to the domain of computer vision such as fine-tuning text-to-image diffusion models using human feedback~\citep{fan2024reinforcement, black2024training}.
and \citet{shankar2024validates} demonstrates how lightweight binary feedback (e.g., thumbs up/down) can effectively guide model behavior.
Similar approaches have been explored in vision and multi-modal research~\citep{fan2024reinforcement, black2024training, le2022deep, pinto2023tuning}. Recently, Reinforcement Learning with CLIP Feedback (RLCF)~\cite{zhao2023test} has been proposed to adapt vision-language models. RLCF relies on the pre-trained CLIP model as a reward function, which may not be available or suitable for all domains or tasks. In contrast, our approach provides a more general and flexible approach for test-time adaptation by effectively guiding the adaptation without relying on specific pre-trained models.

\vspace{-0.35cm}\paragraph{Active learning.}
Active learning~\citep{cohn1994active, settles2009active} involves an oracle (e.g., human annotator) in the machine learning process to enable efficient annotation and training. The active learning framework has been widely studied in active (source-free) domain adaptation~\citep{ash2019deep, clue, elpt, mhpl, du2024diffusion, ning2021multi} and active TTA~\citep{atta}. Compared with active domain adaptation, active TTA focuses on the non-regrettable active sample selection on the continuously changing data stream without access to source data.
Using binary feedback is related to the active learning with partial feedback  problem~\citep{hu2018active}, which seeks to recursively obtain partial labels until a definitive label is identified. \citet{joshi2010breaking} proposed an active learning setup where users compare two images and report whether they belong to the same category. 
% As discussed, existing active labeling methods often require recursive binary questions or diverse data comparisons to tackle the initial low accuracy of the active learning setup. 
In contrast, our approach leverages single-step binary feedback on the model's current batch sample output without requiring additional data. This simplifies the process and reduces the labeling effort.

\section{Conclusion}\label{sec:conclusion}

We proposed test-time adaptation with binary feedback to address the challenge of adapting pre-trained models to new domains with minimal labeling effort. Our approach leverages binary feedback on the model predictions (\textit{correct} or \textit{incorrect}) from an oracle to guide the adaptation process, significantly reducing the labeling cost compared to methods that require full-class labels. 
Our method, \method{}, uniquely combines binary feedback-guided adaptation on uncertain samples with agreement-based self-adaptation on confident samples in a reinforcement learning framework, balancing between a few labeled samples and many unlabeled samples. Through extensive experiments on distribution shift datasets, we demonstrated that \method{} outperforms state-of-the-art test-time adaptation methods, showcasing its effectiveness in handling continuous distribution shifts. 
% By integrating Monte Carlo dropout, our method improves the robustness of model predictions and provides better-calibrated class probability estimates, which are crucial for a triadic learning scheme. 
% The experimental results on distribution shift datasets demonstrate the effectiveness of BATTA. 
%  The experimental results on domain generalization datasets such as PACS and VLCS demonstrate the effectiveness of BATTA. Our approach not only maintains competitive performance but also reduces the labeling burden, making it suitable for real-world applications where extensive labeled data is impractical to obtain. The incorporation of MC dropout further enhances the model's ability to handle uncertain and noisy predictions, contributing to its overall robustness.
Overall, test-time adaptation with binary feedback represents a significant step forward in test-time adaptation, offering a practical balance between performance and labeling efficiency.

% \input{07_ack}

% \input{07_discussion}


\section*{Acknowledgements}
This work was supported by Institute of Information \& communications Technology Planning \& Evaluation (IITP) grant funded by the Korea government (MSIT) (No. RS-2025-02215122, Development and Demonstration of Lightweight AI Model for Smart Homes),
Institute of Information \& communications Technology Planning \& Evaluation (IITP) grant funded by the Korea government (MSIT) (No. RS-2025-02263169, Detection and Prediction of Emerging and Undiscovered Voice Phishing), and
Institute of Information \& communications Technology Planning \& Evaluation(IITP) grant funded by the Korea government(MSIT)
(No. RS-2020-II201336, Artificial Intelligence Graduate School Program(UNIST)).

\section*{Impact Statement}
\system{} and \method{} offer significant societal and practical benefits while presenting some important considerations for implementation. The use of binary feedback substantially reduces labeling costs, making active test-time adaptation more accessible to the end-users. This enables real-time adaptation in critical applications like autonomous driving, healthcare diagnostics, and medical applications; thereby improving safety, efficiency, and user experience. While the system enhances model performance in diverse and changing environments with minimal labeling effort, it requires careful consideration of potential risks, including privacy concerns in surveillance applications and biased decision-making in applications like targeted advertising. 
% With appropriate usages, \method{}'s overall impact appears positive, particularly in democratizing access to adaptive AI systems while maintaining practical efficiency.

% \section*{Reproducibility Statement}
% We provide the source code in the supplementary material with the instructions to prepare the dataset. We specify the experimental details in Appendix~\ref{app:exp_detail}, including datasets, scenarios, and hyperparameters.


\bibliography{references}
\bibliographystyle{icml2025}

%%%%%%%%%%%%%%%%%%%%%%%%%%%%%%%%%%%%%%%%%%%%%%%%%%%%%%%%%%%%%%%%%%%%%%%%%%%%%%%
%%%%%%%%%%%%%%%%%%%%%%%%%%%%%%%%%%%%%%%%%%%%%%%%%%%%%%%%%%%%%%%%%%%%%%%%%%%%%%%
% APPENDIX
%%%%%%%%%%%%%%%%%%%%%%%%%%%%%%%%%%%%%%%%%%%%%%%%%%%%%%%%%%%%%%%%%%%%%%%%%%%%%%%
%%%%%%%%%%%%%%%%%%%%%%%%%%%%%%%%%%%%%%%%%%%%%%%%%%%%%%%%%%%%%%%%%%%%%%%%%%%%%%%
\newpage
\appendix
\onecolumn
\clearpage

\appendix

% \begin{center}
%     \vspace{0.15in}
%     {\bf {\LARGE Appendix}} \\
%     \vspace{0.15in}
%     % {\bf {\Large Test-Time Adaptation with Binary Feedback}}
% \end{center}

%%%%%%%%%%%%%%%%%%%%%%%%%%%%%%%%%%%%%%%%%%%%%%%%%%%%%%%%%%%%

% \newtheorem{theorem}{Theorem}

\section{Discussion}\label{app:impact}

% \subsection{Limitations and future works}

Despite the promising results, the \system{} setting and \method{} method have limitations. First, the reliance on binary feedback, while reducing labeling effort, may still require substantial oracle involvement in scenarios with high data variability or rapid domain shifts. Although \method{} robustly performs across various labeling scenarios (Figures~\ref{fig:ablation_acc_n}, \ref{fig:acc_skip}, and \ref{fig:acc_delay}), future work may explore reducing oracle involvement by developing more advanced and dynamic sample selection strategies. Second, the computational overhead introduced by Monte Carlo dropout (Table~\ref{tab:wall_clock_time}) could be further reduced by efficient TTA~\citep{hong2023mecta, song2023ecotta} and on-device machine learning~\citep{liberis2023differentiable, rusci2020memory, gong2023collaborative}. Finally, although our algorithm robustly outperformed with feedback errors (Figure~\ref{fig:feedback_error}), designing a method for specifically handling noisy or incorrect feedback remains an area for future research.

\section{Additional experiments}\label{app:ablation}

% \vspace{-0.15cm}\paragraph{Using LLMs for labeling.} With recent advances in foundation models, one might argue the use of foundational models for replacing human feedback. We examine the performance of a state-of-the-art foundational model for providing feedback on test-time adaptation. In addition, foundational models perform near-random prediction for unseen tasks such as human sensing tasks~\cite{yoon2024my}. Therefore, incorporating foundation models for (binary-feedback) active TTA is likely to harm the adaptation performance for end-user applications.

% \begin{figure*}[ht]
%     \centering
%     \includegraphics[width=\linewidth]{fig/plot_gpt.pdf}
%     \caption{Accuracy (\%) comparison of foundational model with unsupervised TTA methods in CIFAR10-C.}~\label{fig:gpt}
%     % \vspace{-0.3cm}
%     % \hfill
% \end{figure*}

% \begin{figure}[h]
%     \centering
%     \vspace{-0.5cm}
%     \begin{subfigure}[t]{0.35\linewidth}
%         \centering
%         \includegraphics[width=\linewidth]{fig/conf_threshold.pdf}
%     \end{subfigure}
%     \caption{Accuracy (\%) by confidence thresholding strategies: Ours and hard thresholding (th). Averaged over three random seeds.}
%     \label{fig:conf_threshold}
% \end{figure}

\

\begin{figure}[ht]
    \centering
    \begin{minipage}[h]{0.30\textwidth}
        \centering
        % \vspace{-0.3cm}
        \vspace{0pt}
        \includegraphics[width=\linewidth]{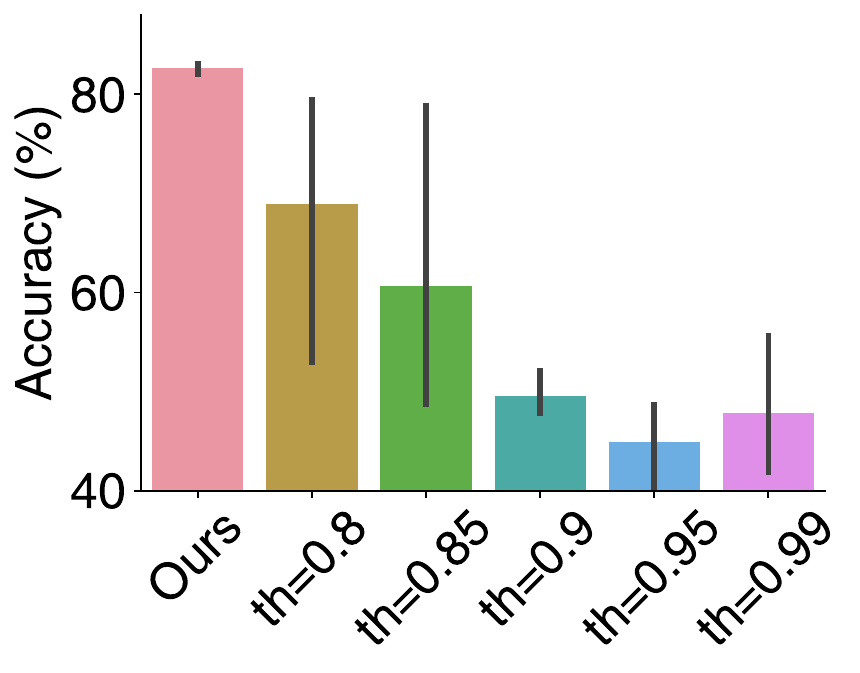}
        % \vspace{-0.3cm}
        \caption{Accuracy (\%) comparison with hard thresholding (th). Ours dynamically selects confident samples via agreement between the model and MC-dropout. Averaged over three random seeds.}    \label{fig:conf_threshold}
    \end{minipage}
    \hfill
    \begin{minipage}[h]{0.30\textwidth}
        \centering
        % \vspace{0.05cm}
        \vspace{0pt}
        \includegraphics[width=\linewidth]{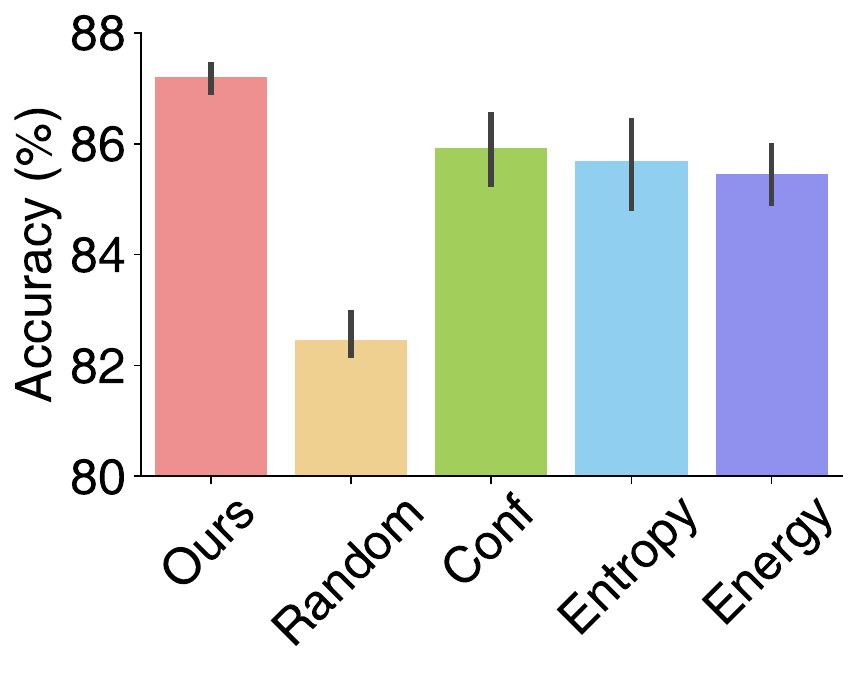}
        % \vspace{-0.05cm}
        \caption{Accuracy (\%) varying the binary-feedback sample selection strategy in CIFAR10-C. Ours leverages MC-dropout to select the most uncertain samples. Averaged over three random seeds.}
        \label{fig:sample_selection}
    \end{minipage}
    \hfill
    \begin{minipage}[h]{0.30\textwidth}
        \centering
        % \vspace{-0.00cm}
        % \vspace{2pt}
        \vspace{0cm}
        \includegraphics[width=\linewidth]{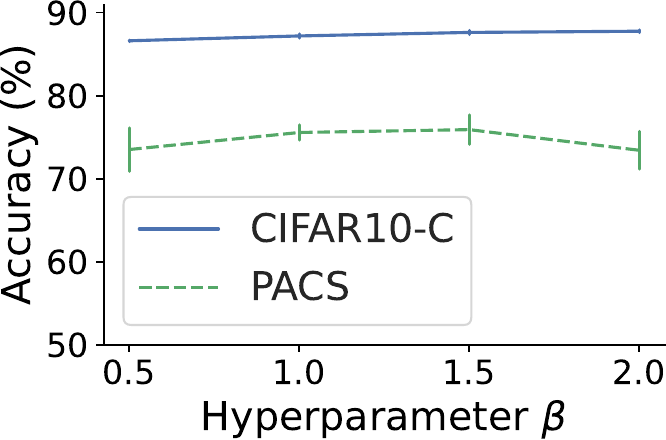}
        \vspace{-0.2cm}
        \caption{Accuracy (\%) varying the balancing hyperparameter ($\beta$) in CIFAR10-C and PACS. Another balancing hyperparameter $\alpha$ is set to 1. Averaged over three random seeds.}
        \label{fig:ablation_lambda}
    \end{minipage}
\end{figure}
\begin{figure}[ht]
    \begin{minipage}[t]{0.30\textwidth}
        \centering
        % \vspace{-0.1cm}
        % \vspace{1.7pt}
        \includegraphics[width=\linewidth]{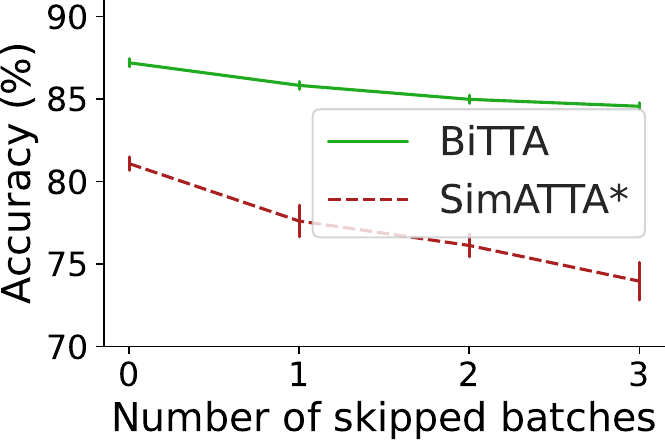}
        % \vspace{-0.15cm}
        % \vspace{0.3pt}
        \caption{Accuracy (\%) varying the labeling skip in CIFAR10-C. Averaged over three random seeds.}
        \label{fig:acc_skip}
    \end{minipage}
    \hfill
    \begin{minipage}[t]{0.30\textwidth}
        \centering
        % \vspace{-0.1cm}
        % \vspace{1.7pt}
        \includegraphics[width=\linewidth]{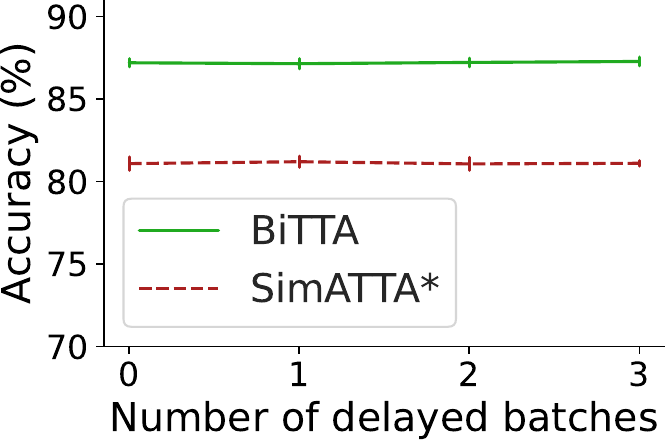}
        % \vspace{-0.15cm}
        % \vspace{0.3pt}
        \caption{Accuracy (\%) varying the labeling delay (labeled samples arrive after certain batches). Averaged over three random seeds.}
        \label{fig:acc_delay}
    \end{minipage}
    \hfill
    \begin{minipage}[t]{0.30\textwidth}
        \centering
        % \vspace{-0.1cm}
        % \vspace{1.7pt}
        \includegraphics[width=\linewidth]{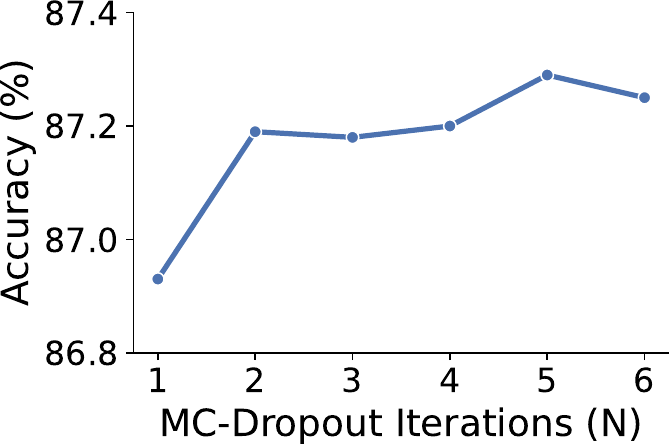}
        % \vspace{-0.15cm}
        % \vspace{0.3pt}
        \caption{Accuracy (\%) varying the number of dropouts in MC-dropout. Averaged over three random seeds.}
        \label{fig:mc_dropout_ablation}
    \end{minipage}
\end{figure}

\vspace{-0.15cm}\paragraph{Comparing prediction agreement with confidence thresholding.} To assess the effectiveness of our prediction agreement method for confident sample selection, we compared it against fixed confidence thresholding approaches. We evaluated thresholds ranging from 0.8 to 0.99, with 0.99 being the value used in SoTTA~\citep{sotta}. Figure~\ref{fig:conf_threshold} illustrates the performance of these approaches on unlabeled-only TTA in the continual CIFAR10-C setting. Our prediction agreement method consistently outperformed all fixed thresholding approaches, which exhibited high variance and instability. This result demonstrates the superiority of our dynamic sample selection strategy, particularly in scenarios with continuously changing corruptions, highlighting the importance of adaptive confidence assessment in test-time adaptation.

\vspace{-0.15cm}\paragraph{Impact of sample selection.}
We examined the impact of sample selection, including our MC-dropout certainty approach with random selection, maximum entropy~\citep{saito2020universal}, minimum confidence~\citep{sohn2020fixmatch}, and minimum energy~\citep{ood_energy}. In Figure~\ref{fig:sample_selection}, our method outperforms others by leveraging MC-dropout to estimate epistemic uncertainty. In contrast, naive methods may struggle with overconfidence in test-time scenarios, failing to prioritize samples that offer the most valuable information for model improvement. On the other hand, we observe that for large datasets (e.g., ImageNet-C/R and VisDA-2021 in Table~\ref{tab:additional_datasets}), most predictions are already unconfident and incorrect, so random selection is sufficient to capture uncertain samples for effective adaptation.

% \begin{figure}[h]
%     \centering
%     \begin{subfigure}[t]{0.35\linewidth}
%         \centering
%         \includegraphics[width=\linewidth]{fig/accuracy_lambda.pdf}
%     \end{subfigure}
%     \caption{Accuracy (\%) varying the balancing hyperparameter ($\beta$) in CIFAR10-C and PACS. Averaged over three random seeds.}
%     \label{fig:ablation_lambda}
% \end{figure}

\vspace{-0.15cm}\paragraph{Sensitivity to balancing hyperparameter $\alpha, \beta$.} We investigated the sensitivity of \method{} to the balancing hyperparameter $\beta$ while fixing $\alpha=1.0$, which controls the trade-off between binary feedback-guided adaptation and agreement-based self-adaptation. Figure~\ref{fig:ablation_lambda} illustrates the overall accuracy across various $\beta$ values for both image corruption and domain adaptation datasets. The results demonstrate that \method{} maintains consistent performance across a wide range of $\beta$ values, indicating robustness to this hyperparameter choice. This stability suggests that \method{} can effectively deploy across different scenarios without extensive hyperparameter tuning.

\vspace{-0.15cm}\paragraph{Impact of intermittent and delayed labeling.} To further understand the impact of the annotator's labeling budget, we conduct an experiment scenario where annotators skip or delay the labeling of a few batches (e.g., labeling only 1 out of 4 consecutive batches). In Figures~\ref{fig:acc_skip} and \ref{fig:acc_delay}, we observe our \method{} shows stable performance with minimal degradation, whereas the active TTA baseline (SimATTA) shows high accuracy degradation with batch skips and consistent suboptimal performance with delayed batches. The result supports the robustness of \method{} in varying labeling scenarios, enabling a practical and scalable TTA.

\vspace{-0.15cm}\paragraph{Runtime analysis.} To assess the practical applicability of \method{}, we conducted a comprehensive runtime analysis by measuring the average wall-clock time per batch across different methods on the Tiny-ImageNet-C dataset. 
Our results in Table~\ref{tab:wall_clock_time} show that \method{} requires 4.19 ±0.06 seconds per batch, positioning it between simpler TTA methods (0.33-1.68s) and more complex approaches like CoTTA (26.63s) and SimATTA (45.45s).
The runtime profile demonstrates that \method{} achieves a favorable balance between computational cost and performance, particularly considering its significant accuracy improvements over faster baselines while maintaining substantially lower processing time than methods like SimATTA.

\begin{table*}[t]
\centering
% \captionsetup{labelfont={color=blue}}
\caption{Average wall-clock time per batch (s) comparisons with TTA and active TTA baselines with binary feedback in Tiny-ImageNet-C. Notation * indicates the modified algorithm to utilize binary-feedback samples. Averaged over three random seeds.}
\label{tab:wall_clock_time}

\footnotesize

\resizebox{\linewidth}{!}{%
\begin{tabularx}{1.35\textwidth}{l *{9}Y
>{\columncolor[HTML]{DCFFDC}}Y }
\toprule
 & SrcValid & BN-Stats & TENT* & EATA* & SAR* & CoTTA* & RoTTA* & SoTTA* & SimATTA* & \method{} \\
 \midrule
Avg. & 0.18 \scalebox{\std}{±0.12} & 0.33 \scalebox{\std}{±0.20} & 1.03 \scalebox{\std}{±0.35} & 0.98 \scalebox{\std}{±0.39} & 1.02 \scalebox{\std}{±0.38} & 26.63 \scalebox{\std}{±5.40} & 1.68 \scalebox{\std}{±0.27} & 1.25 \scalebox{\std}{±0.16} & 45.45 \scalebox{0.6}{±13.50} & 4.19 \scalebox{\std}{±0.06}\\
\bottomrule
\end{tabularx}
}

\end{table*}

\vspace{-0.15cm}\paragraph{Memory analysis.}
We report the average GPU memory consumption across all CIFAR10-C corruptions (severity 5) in Table~\ref{tab:gpu_memory}. Compared to baseline methods, \method{} shows higher memory usage due to repeated forward passes for MC-dropout and the reinforcement learning formulation. To mitigate this, we evaluate two variants: {\method{}+MECTA} (applying the memory-efficient continual test-time adaptation from MECTA~\cite{hong2023mecta}) and {\method{}+GA} (using gradient accumulation to divide a batch into multiple mini-batches). \method{}+GA reduces memory usage up to $60\%$ while retaining \method{}’s performance, demonstrating a practical strategy for deployment in memory-constrained settings.

\begin{table*}[t]
\centering
\caption{Average GPU memory consumption (MB). MECTA and gradient accumulation (GA) are applied to \method{}. Notation * indicates the modified algorithm to utilize binary-feedback samples.}
\label{tab:gpu_memory}

\footnotesize

\resizebox{\linewidth}{!}{%
\begin{tabularx}{1.35\textwidth}{l *{9}Y >{\columncolor[HTML]{DCFFDC}}Y >{\columncolor[HTML]{DCFFDC}}Y >{\columncolor[HTML]{DCFFDC}}Y}
\toprule
 & SrcValid & BN-Stats & TENT* & EATA* & SAR* & CoTTA* & RoTTA* & SoTTA* & SimATTA* & \method{} & +MECTA & +GA \\
\midrule
Avg.& 2081 & 2696 & 3246 & 3239 & 3244 & 2966 & 3038 & 3229 & 2824 & 8304 & 6724 & {2841} \\
\bottomrule
\end{tabularx}
}

\end{table*}

\vspace{-0.15cm}\paragraph{Impact of the number of epochs.}
To understand the \method{}'s performance under time-constrained environments, we examined how reducing training epochs affects adaptation accuracy on CIFAR10-C. We compared our standard 3-epoch configuration against reduced 1- and 2-epoch settings, adjusting learning rates proportionally (×3 and ×1.5) to compensate for fewer update steps. Results in Table~\ref{tab:app_epoch} show that \method{} maintains robust performance even with fewer epochs. This consistent performance across epoch configurations demonstrates that \method{} can effectively adapt to distribution shifts even under stricter computational constraints, offering flexibility in real-world deployment scenarios where faster adaptation may be preferred.

\begin{table}[t]
\centering
% \captionsetup{labelfont={color=blue}}
\caption{Accuracy (\%) comparisons with varying epochs in CIFAR10-C (severity level 5).  B: TTA with binary feedback.   Averaged over three random seeds.}
\label{tab:app_epoch}
\footnotesize

\resizebox{\columnwidth}{!}{%
\begin{tabularx}{1.4\textwidth}{c l *{16}Y}
\toprule
& & \multicolumn{3}{c}{Noise} & \multicolumn{4}{c}{Blur} & \multicolumn{4}{c}{Weather} & \multicolumn{4}{c}{Digital} &  \\
& & \multicolumn{15}{l}{ %$t$ 
    \begin{tikzpicture}[overlay, remember picture]
      \draw[->, line width=0.2pt] (0,0.075) -- (19.7,0.075);
    \end{tikzpicture}
    } &  \\
\multirow{-3}{*}{Label} & \multirow{-3}{*}{Method} & Gau. & Shot & Imp. & Def. & Gla. & Mot. & Zoom & Snow & Fro. & Fog & Brit. & Cont. & Elas. & Pix. & JPEG & Avg. \\
\midrule
B & \cellcolor[HTML]{DCFFDC}\method{} (epoch = 3) & \cellcolor[HTML]{DCFFDC}76.78 & \cellcolor[HTML]{DCFFDC}84.24 & \cellcolor[HTML]{DCFFDC}78.75 & \cellcolor[HTML]{DCFFDC}87.51 & \cellcolor[HTML]{DCFFDC}77.39 & \cellcolor[HTML]{DCFFDC}88.38 & \cellcolor[HTML]{DCFFDC}91.36 & \cellcolor[HTML]{DCFFDC}89.42 & \cellcolor[HTML]{DCFFDC}90.72 & \cellcolor[HTML]{DCFFDC}90.30 & \cellcolor[HTML]{DCFFDC}94.65 & \cellcolor[HTML]{DCFFDC}92.62 & \cellcolor[HTML]{DCFFDC}86.15 & \cellcolor[HTML]{DCFFDC}92.42 & \cellcolor[HTML]{DCFFDC}87.24 & \cellcolor[HTML]{DCFFDC}\textbf{87.20} \\
B & \cellcolor[HTML]{DCFFDC}$\cdot$ epoch = 1 & \cellcolor[HTML]{DCFFDC}76.92 & \cellcolor[HTML]{DCFFDC}84.29 & \cellcolor[HTML]{DCFFDC}78.61 & \cellcolor[HTML]{DCFFDC}86.99 & \cellcolor[HTML]{DCFFDC}77.20 & \cellcolor[HTML]{DCFFDC}88.36 & \cellcolor[HTML]{DCFFDC}91.51 & \cellcolor[HTML]{DCFFDC}89.31 & \cellcolor[HTML]{DCFFDC}90.58 & \cellcolor[HTML]{DCFFDC}90.30 & \cellcolor[HTML]{DCFFDC}94.51 & \cellcolor[HTML]{DCFFDC}92.70 & \cellcolor[HTML]{DCFFDC}85.77 & \cellcolor[HTML]{DCFFDC}92.08 & \cellcolor[HTML]{DCFFDC}87.50 & \cellcolor[HTML]{DCFFDC}87.11 \\
B & \cellcolor[HTML]{DCFFDC}$\cdot$ epoch = 2 & \cellcolor[HTML]{DCFFDC}76.30 & \cellcolor[HTML]{DCFFDC}84.01 & \cellcolor[HTML]{DCFFDC}78.80 & \cellcolor[HTML]{DCFFDC}87.66 & \cellcolor[HTML]{DCFFDC}77.30 & \cellcolor[HTML]{DCFFDC}88.43 & \cellcolor[HTML]{DCFFDC}91.56 & \cellcolor[HTML]{DCFFDC}89.16 & \cellcolor[HTML]{DCFFDC}90.61 & \cellcolor[HTML]{DCFFDC}90.37 & \cellcolor[HTML]{DCFFDC}94.52 & \cellcolor[HTML]{DCFFDC}92.61 & \cellcolor[HTML]{DCFFDC}85.83 & \cellcolor[HTML]{DCFFDC}92.33 & \cellcolor[HTML]{DCFFDC}87.75 & \cellcolor[HTML]{DCFFDC}87.15
 \\
\bottomrule
\end{tabularx}
}
\end{table}

\vspace{-0.15cm}\paragraph{MC-dropout configuration.}
We used 4 dropout inferences ($N=4$) for policy estimation in \method{}. To evaluate sensitivity to the number of MC-dropout inferences, we conducted an ablation study varying $N$ from 1 to 6. As shown in Figure~\ref{fig:mc_dropout_ablation}, \method{} performs consistently well for all $N > 1$, demonstrating robustness to this hyperparameter.
Even at $N=1$, MC-dropout is still active, introducing stochasticity essential for both core components: (1) {BFA} relies on uncertainty to select feedback samples, and (2) {ABA} leverages agreement between stochastic and deterministic predictions. In contrast, removing dropout entirely leads to a $2.56\%$ accuracy drop.
Although gains from increasing $N$ are modest, higher $N$ improves uncertainty calibration, reducing the expected calibration error (ECE) from $0.142$ at $N=1$ to $0.064$ at $N=4$. We thus adopt $N=4$ as a balanced choice. In practice, smaller $N$ may be preferred for low-latency scenarios, while larger $N$ may benefit reliability-critical deployments.

\vspace{-0.15cm}\paragraph{Replacing MC-dropout uncertainty estimation.}
To further understand the importance of MC-dropout uncertainty estimation, we compare \method{} with replacing MC-dropout with augmentation-based estimation (as in \citet{cotta, memo}), ensemble-based estimation~\cite{tast}, and simple deterministic softmax confidence. Results in Table~\ref{tab:app_augment} suggest that augmentation-based uncertainty appears less stable and overfits in the early adaptation stage, leading to suboptimal performance.

\begin{table}[t]
\centering
% \captionsetup{labelfont={color=blue}}
\caption{Accuracy (\%) comparisons with replacing MC-dropout uncertainty estimation with (1) sample augmentation, (2) ensemble method, and (3) original confidence in CIFAR10-C (severity level 5).  B: TTA with binary feedback.   Averaged over three random seeds.}
\label{tab:app_augment}
\footnotesize

\resizebox{\columnwidth}{!}{%
\begin{tabularx}{1.4\textwidth}{c l *{16}Y}
\toprule
& & \multicolumn{3}{c}{Noise} & \multicolumn{4}{c}{Blur} & \multicolumn{4}{c}{Weather} & \multicolumn{4}{c}{Digital} &  \\
& & \multicolumn{15}{l}{ %$t$ 
    \begin{tikzpicture}[overlay, remember picture]
      \draw[->, line width=0.2pt] (0,0.075) -- (20.25,0.075);
    \end{tikzpicture}
    } &  \\
\multirow{-3}{*}{Label} & \multirow{-3}{*}{Method} & Gau. & Shot & Imp. & Def. & Gla. & Mot. & Zoom & Snow & Fro. & Fog & Brit. & Cont. & Elas. & Pix. & JPEG & Avg. \\
\midrule
B & \cellcolor[HTML]{DCFFDC}\method{} & \cellcolor[HTML]{DCFFDC}76.78 & \cellcolor[HTML]{DCFFDC}84.24 & \cellcolor[HTML]{DCFFDC}78.75 & \cellcolor[HTML]{DCFFDC}87.51 & \cellcolor[HTML]{DCFFDC}77.39 & \cellcolor[HTML]{DCFFDC}88.38 & \cellcolor[HTML]{DCFFDC}91.36 & \cellcolor[HTML]{DCFFDC}89.42 & \cellcolor[HTML]{DCFFDC}90.72 & \cellcolor[HTML]{DCFFDC}90.30 & \cellcolor[HTML]{DCFFDC}94.65 & \cellcolor[HTML]{DCFFDC}92.62 & \cellcolor[HTML]{DCFFDC}86.15 & \cellcolor[HTML]{DCFFDC}92.42 & \cellcolor[HTML]{DCFFDC}87.24 & \cellcolor[HTML]{DCFFDC}\textbf{87.20} \\
B & $\cdot$ Augmentation & 66.22 & 46.99 & 25.43 & 18.49 & 12.82 & 11.96 & 11.68 & 11.43 & 12.24 & 11.37 & 11.48 & 10.87 & 11.45 & 11.96 & 11.71 & 19.07
 \\
B & $\cdot$ Ensemble & 74.08 & 81.60 & 75.57 & 88.16 & 74.15 & 88.50 & 91.02 & 87.93 & 89.71 & 88.92 & 94.05 & 92.01 & 83.28 & 89.83 & 83.64 & 85.50
\\
B & $\cdot$ Confidence & 74.00 & 82.61 & 76.54 & 87.12 & 75.13 & 87.83 & 90.92 & 88.14 & 89.90 & 89.20 & 94.33 & 92.28 & 84.08 & 90.93 & 85.54 & 85.90
\\
\bottomrule
\end{tabularx}
}
\end{table}

\section{Additional results}\label{app:add_result}

\paragraph{Results on additional datasets.}
We conduct an additional experiment to evaluate the scalability of \method{} across various datasets covered in recent works~\citep{deyo, sar, atta, chen2022contrastive}: ImageNet-C~\citep{cifarc}, ImageNet-R~\citep{imagenetr}, ColoredMNIST~\citep{arjovsky2019invariant}, VisDA-2021~\citep{visda2021}, and DomainNet~\citep{domainnet}. Results in Table~\ref{tab:additional_datasets} demonstrate a superior performance of \method{}, especially on large-scale datasets such as ImageNet-C. The key insight is that \method{} formulates both binary feedback and unlabeled sample adaptation as a single reinforcement learning objective, where the reward signals seamlessly guide the model's adaptation. Also, the use of MC-dropout provides a robust uncertainty estimate, while optimizing on MC-dropout prevents the TTA model from overfitting, therefore showing a stable adaptation in large-scale datasets.

% \begin{table}[t]
% \centering
% \caption{Accuracy (\%) comparisons with varying MC-dropout inferences ($N$) on CIFAR10-C. Averaged over three random seeds.}

% \scriptsize
% \begin{tabularx}{0.5\textwidth}{lYYYYYY}
% \toprule
% $N$ & 1 & 2 & 3 & 4 & 5 & 6 \\
% \midrule
% \method & 86.93 & 87.19 & 87.18 & 87.20 & {87.29} & 87.25 \\
% \bottomrule
% \end{tabularx}
% \label{tab:mc_dropout_ablation}
% \end{table}

\begin{table*}[t]
\centering
% \captionsetup{labelfont={color=blue}}
\caption{Accuracy (\%) comparisons with TTA and active TTA baselines in additional datasets. Notation * indicates the modified algorithm to utilize binary-feedback samples. Results outperforming all other baselines are highlighted in \textbf{bold} fonts. Averaged over three random seeds.}
\label{tab:additional_datasets}

\footnotesize

\resizebox{\linewidth}{!}{%
\begin{tabularx}{1.25\textwidth}{l *{9}Y
>{\columncolor[HTML]{DCFFDC}}Y }
\toprule
Dataset & SrcValid & BN-Stats & TENT* & EATA* & SAR* & CoTTA* & RoTTA* & SoTTA* & SimATTA* & \method{} \\
 \midrule
ImageNet-C & 14.43 & 26.88 & 0.93 & 30.87 & 35.15 & 26.80 & 22.55 & 36.02 & 17.50 & \textbf{36.59} \\[2pt]
ImageNet-R & 33.05 & 35.08 & 29.10 & 37.14 & 36.64 & 35.02 & 34.35 & 31.00 & 35.63 & \textbf{38.59} \\[2pt]
VisDA-2021 & 27.36 & 26.46 & 20.38 & 27.82 & 27.41 & 26.46 & 27.23 & 27.71 & 22.80 & \textbf{29.30} \\[2pt]
DomainNet & 54.82 & 54.41 & 18.80 & 59.49 & 57.78 & 54.40 & 56.41 & 54.82 & 58.41 & \textbf{60.85} \\[2pt]
ColoredMNIST & 50.49 & 45.59 & 44.92 & 45.59 & 45.74 & 45.60 & 48.90 & 59.45 & 93.66 & \textbf{96.75}
\\
\bottomrule
\end{tabularx}
}

\end{table*}

\vspace{-0.15cm}\paragraph{Results on additional scenarios.}
Recent TTA works suggest a new scenario of (1) imbalanced/non-iid label distribution, where ground-truth labels are temporally correlated~\citep{sar, note}, (2) and batch size 1~\citep{sar}. Note that SimATTA's clustering algorithm for sample selection is not applicable in scenarios where the memory capacity is limited to only one image. Experiment results on CIFAR10-C (Table~\ref{tab:app_noniid}) suggest the robustness of our method over imbalanced label distribution and batch size 1 by effectively utilizing reward signals from the binary feedback and unlabeled samples.

\begin{table}[t]
\centering
% \captionsetup{labelfont={color=blue}}
\caption{Accuracy (\%) comparisons with TTA and active TTA baselines with binary feedback in online CIFAR10-C (severity level 5) with additional scenarios. Notation * indicates the modified algorithm to utilize binary-feedback samples. B: TTA with binary feedback.  Results outperforming all other baselines are highlighted in \textbf{bold} fonts. Averaged over three random seeds.}
\label{tab:app_noniid}
\footnotesize

\begin{subtable}{\textwidth}
\resizebox{\columnwidth}{!}{%
\begin{tabularx}{1.3\textwidth}{c l *{16}Y}

\toprule
& & \multicolumn{3}{c}{Noise} & \multicolumn{4}{c}{Blur} & \multicolumn{4}{c}{Weather} & \multicolumn{4}{c}{Digital} & \\
\cmidrule(lr){3-18}
\multirow{-2}{*}{Label} & \multirow{-2}{*}{Method} & Gau. & Shot & Imp. & Def. & Gla. & Mot. & Zoom & Snow & Fro. & Fog & Brit. & Cont. & Elas. & Pix. & JPEG & Avg. \\
\midrule
- & SrcValid & 24.01 & 30.91 & 22.36 & 55.00 & 53.44 & 66.99 & 63.74 & 78.01 & 68.41 & 73.92 & 91.34 & 34.30 & 76.77 & 46.26 & 73.05 & 57.70 \\
- & BN-Stats & 22.75 & 23.33 & 20.83 & 30.15 & 21.45 & 29.38 & 28.90 & 27.33 & 28.05 & 29.27 & 31.37 & 31.06 & 25.21 & 26.37 & 22.91 & 26.58 \\
B & TENT* & 20.00 & 21.27 & 19.56 & 26.77 & 19.19 & 26.54 & 25.76 & 24.94 & 24.66 & 26.50 & 28.03 & 26.66 & 22.14 & 23.88 & 20.98 & 23.79 \\  
B & EATA* & 16.24 & 16.52 & 13.73 & 18.82 & 15.97 & 18.87 & 18.79 & 16.87 & 17.62 & 19.30 & 20.34 & 17.85 & 18.02 & 17.87 & 16.29 & 17.54 \\
B & SAR* & 22.95 & 23.57 & 21.36 & 30.06 & 21.44 & 29.52 & 28.81 & 27.38 & 28.10 & 29.48 & 31.40 & 30.69 & 24.90 & 26.46 & 23.31 & 26.63 \\
B & CoTTA* & 22.76 & 23.36 & 21.14 & 29.99 & 21.42 & 29.48 & 28.84 & 27.42 & 28.10 & 29.43 & 31.34 & 30.85 & 24.92 & 26.43 & 23.22 & 26.58 \\
B & RoTTA* & 41.83 & 44.60 & 37.97 & 58.54 & 41.14 & 57.40 & 57.79 & 52.54 & 51.86 & 56.87 & 62.27 & 53.20 & 48.41 & 50.65 & 44.84 & 50.66 \\
B & SoTTA* & 67.03 & 71.31 & 61.84 & 83.96 & 66.01 & 82.23 & 84.47 & 78.62 & 78.48 & 82.94 & 87.74 & 77.29 & 74.07 & 76.94 & 72.12 & 76.34 \\
B & SimATTA* & 59.05 & 68.67 & 44.43 & 84.96 & 67.46 & 83.36 & 84.99 & 81.75 & 82.87 & 83.83 & 89.11 & 72.28 & 76.15 & 81.90 & 73.41 & 75.62 \\
B & \cellcolor[HTML]{DCFFDC}\method{} & \cellcolor[HTML]{DCFFDC}\textbf{82.32} & \cellcolor[HTML]{DCFFDC}\textbf{84.02} & \cellcolor[HTML]{DCFFDC}\textbf{75.77} & \cellcolor[HTML]{DCFFDC}\textbf{90.39} & \cellcolor[HTML]{DCFFDC}\textbf{79.05} & \cellcolor[HTML]{DCFFDC}\textbf{90.73} & \cellcolor[HTML]{DCFFDC}\textbf{90.93} & \cellcolor[HTML]{DCFFDC}\textbf{90.71} & \cellcolor[HTML]{DCFFDC}\textbf{89.09} & \cellcolor[HTML]{DCFFDC}\textbf{92.22} & \cellcolor[HTML]{DCFFDC}\textbf{95.36} & \cellcolor[HTML]{DCFFDC}\textbf{82.16} & \cellcolor[HTML]{DCFFDC}\textbf{87.56} & \cellcolor[HTML]{DCFFDC}\textbf{87.40} & \cellcolor[HTML]{DCFFDC}\textbf{85.91} & \cellcolor[HTML]{DCFFDC}\textbf{86.91}
\\
\bottomrule
\end{tabularx}
}
\vspace{0.05cm}
\caption{Imbalanced (non-iid) label distribution.}
\end{subtable}

\begin{subtable}{\textwidth}
\resizebox{\columnwidth}{!}{%
\begin{tabularx}{1.3\textwidth}{c l *{16}Y}

\toprule
& & \multicolumn{3}{c}{Noise} & \multicolumn{4}{c}{Blur} & \multicolumn{4}{c}{Weather} & \multicolumn{4}{c}{Digital} &  \\
\cmidrule(lr){3-18}
\multirow{-2}{*}{Label} & \multirow{-2}{*}{Method} & Gau. & Shot & Imp. & Def. & Gla. & Mot. & Zoom & Snow & Fro. & Fog & Brit. & Cont. & Elas. & Pix. & JPEG & Avg. \\
\midrule
- & SrcValid & 25.96 & 33.19 & 24.71 & 56.73 & 52.02 & 67.37 & 64.80 & \textbf{77.97} & 67.00 & 74.14 & \textbf{91.50} & {33.90} & \textbf{76.61} & 46.38 & \textbf{73.23} & 57.70 \\
- & BN-Stats & 20.53 & 21.09 & 18.15 & 32.45 & 20.72 & 33.45 & 30.49 & 28.76 & 29.29 & 33.34 & 36.96 & 40.55 & 24.20 & 25.95 & 21.43 & 27.82 \\
B & TENT* & 10.50 & 10.01 & 10.01 & 10.01 & 10.01 & 10.01 & 10.01 & 10.01 & 10.01 & 10.01 & 10.01 & 10.01 & 10.01 & 10.01 & 10.01 & 10.04 \\
B & EATA* & 20.53 & 21.09 & 18.15 & 32.45 & 20.72 & 33.45 & 30.49 & 28.76 & 29.29 & 33.34 & 36.96 & 40.55 & 24.20 & 25.95 & 21.43 & 27.82 \\
B & SAR* & 20.56 & 21.12 & 18.29 & 32.51 & 20.86 & 33.59 & 30.67 & 29.12 & 29.51 & 33.68 & 37.52 & 41.15 & 24.70 & 26.57 & 21.98 & 28.12 \\
B & CoTTA* & 20.54 & 21.09 & 18.15 & 32.44 & 20.70 & 33.45 & 30.49 & 28.75 & 29.28 & 33.33 & 36.95 & 40.55 & 24.20 & 25.95 & 21.42 & 27.82 \\
B & RoTTA* & 11.70 & 10.23 & 10.03 & 10.01 & 10.01 & 10.01 & 10.01 & 10.01 & 10.01 & 10.01 & 10.01 & 10.01 & 10.01 & 10.01 & 10.01 & 10.14 \\
B & SoTTA* & 17.02 & 15.32 & 13.00 & 79.00 & 18.17 & 57.44 & 63.39 & 51.26 & 49.67 & 61.47 & 64.84 & \textbf{50.27} & 53.56 & 42.18 & 52.14 & 45.92 \\
B & \cellcolor[HTML]{DCFFDC}\method{} & \cellcolor[HTML]{DCFFDC}\textbf{62.14} & \cellcolor[HTML]{DCFFDC}\textbf{64.01} & \cellcolor[HTML]{DCFFDC}\textbf{55.13} & \cellcolor[HTML]{DCFFDC}\textbf{82.07} & \cellcolor[HTML]{DCFFDC}\textbf{59.64} & \cellcolor[HTML]{DCFFDC}\textbf{79.22} & \cellcolor[HTML]{DCFFDC}\textbf{83.26} & \cellcolor[HTML]{DCFFDC}75.84 & \cellcolor[HTML]{DCFFDC}\textbf{71.26} & \cellcolor[HTML]{DCFFDC}\textbf{81.92} & \cellcolor[HTML]{DCFFDC}86.13 & \cellcolor[HTML]{DCFFDC}31.94 & \cellcolor[HTML]{DCFFDC}71.34 & \cellcolor[HTML]{DCFFDC}\textbf{73.80} & \cellcolor[HTML]{DCFFDC}67.73 & \cellcolor[HTML]{DCFFDC}\textbf{70.17}
\\
\bottomrule
\end{tabularx}
}
\vspace{0.05cm}
\caption{Batch size 1.}
\end{subtable}

\end{table}

\begin{table}[t]
    \centering
    \begin{minipage}[t]{0.48\textwidth}
    
\centering
\scriptsize
\caption{Average accuracy (\%) comparisons. Notation * indicates the modified algorithm to utilize binary-feedback samples. Averaged over three random seeds.}
\label{tab:add_baselines}

\begin{tabular}{clccc}
\toprule
Label & Method & C10-C & C100-C & T-IN-C \\
\midrule
B & DeYO*      & 84.41 & 61.30 & 40.67 \\
B & TAST-BN*   & 75.53 & 29.13 & 17.05 \\
B & \cellcolor[HTML]{DCFFDC}\method{}      & \cellcolor[HTML]{DCFFDC}\textbf{87.20} & \cellcolor[HTML]{DCFFDC}\textbf{62.49} & \cellcolor[HTML]{DCFFDC}\textbf{40.85} \\
\bottomrule
\end{tabular}

\end{minipage}
\hfill
\begin{minipage}[t]{0.48\textwidth}

\centering
\scriptsize
\caption{Average accuracy (\%) comparisons under the OWTTT pre-trained model. Notation * indicates the modified algorithm to utilize binary-feedback samples. Averaged over three random seeds.}
\label{tab:add_baselines_owttt}

\begin{tabular}{clcc}
\toprule
Label & Method & C10-C & C100-C \\
\midrule
- & OWTTT      & 54.63 & 29.10 \\
B & OWTTT*     & 31.24 & 3.39 \\
B & \cellcolor[HTML]{DCFFDC}\method{}      & \cellcolor[HTML]{DCFFDC}\textbf{89.89} & \cellcolor[HTML]{DCFFDC}\textbf{64.06} \\
\bottomrule
\end{tabular}

\end{minipage}
\end{table}

\paragraph{Comparison with recent baselines}
We incorporated new comparisons with recent TTA methods: DeYO~\citep{deyo}, OWTTT~\citep{owttt}, and TAST~\citep{tast}. As shown in Tables~\ref{tab:add_baselines} and \ref{tab:add_baselines_owttt}, \method{} consistently outperforms these methods under equal binary feedback conditions.

\vspace{-0.15cm}\paragraph{Results on additional architectures.}
To further examine the applicability of \method{} in various model architectures, we experimented with ResNet50 and ViT-Base. Table~\ref{tab:app_resnet50} shows the overall result, where \method{} still outperformed the baselines in all corruptions, demonstrating the applicability of \method{} over diverse models.

\begin{table}[t]
\centering
\caption{Accuracy (\%) comparisons with TTA and active TTA baselines with binary feedback in CIFAR10-C (severity level 5) with additional architectures. Notation * indicates the modified algorithm to utilize binary-feedback samples. B: TTA with binary feedback.  Results outperforming all other baselines are highlighted in \textbf{bold} fonts. Averaged over three random seeds.}
\label{tab:app_resnet50}
\footnotesize

\begin{subtable}{\textwidth}
\resizebox{\columnwidth}{!}{%
\begin{tabularx}{1.3\textwidth}{c l *{16}Y}
\toprule
& & \multicolumn{3}{c}{Noise} & \multicolumn{4}{c}{Blur} & \multicolumn{4}{c}{Weather} & \multicolumn{4}{c}{Digital} &  \\
& & \multicolumn{15}{l}{ %$t$ 
    \begin{tikzpicture}[overlay, remember picture]
      \draw[->, line width=0.2pt] (0,0.075) -- (19,0.075);
    \end{tikzpicture}
    } &  \\
\multirow{-3}{*}{Label} & \multirow{-3}{*}{Method} & Gau. & Shot & Imp. & Def. & Gla. & Mot. & Zoom & Snow & Fro. & Fog & Brit. & Cont. & Elas. & Pix. & JPEG & Avg. \\
\midrule
- & SrcValid & 22.56 & 27.66 & 21.49 & 46.91 & 43.23 & 55.29 & 54.62 & 66.90 & 53.91 & 61.31 & 84.94 & 24.24 & 65.29 & 41.03 & 65.35 & 48.98 \\
- & BN-Stats & 60.20 & 62.13 & 55.50 & 82.21 & 58.39 & 80.01 & 81.65 & 75.67 & 73.78 & 78.92 & 86.14 & 81.86 & 69.56 & 73.34 & 67.23 & 72.44 \\
B & TENT* & 67.91 & 72.96 & 63.60 & 72.68 & 56.98 & 62.43 & 65.48 & 60.95 & 58.81 & 56.47 & 66.26 & 64.45 & 55.80 & 61.30 & 57.70 & 61.58 \\
B & EATA* & 75.19 & 80.89 & 73.29 & 81.65 & 67.68 & 76.30 & 79.09 & 75.80 & 77.09 & 76.19 & 82.23 & 79.64 & 68.67 & 74.07 & 70.10 & 75.62 \\
B & SAR* & 63.51 & 70.85 & 65.95 & 85.07 & 66.46 & 84.06 & 86.33 & 82.68 & 83.24 & 84.02 & 90.46 & 86.74 & 78.53 & 83.68 & 79.10 & 79.53 \\
B & CoTTA* & 60.20 & 62.13 & 55.50 & 82.20 & 58.40 & 80.01 & 81.65 & 75.68 & 73.78 & 78.92 & 86.14 & 81.87 & 69.55 & 73.36 & 67.20 & 72.63 \\
B & RoTTA* & 60.77 & 65.94 & 60.67 & 79.87 & 65.04 & 82.22 & 86.19 & 82.03 & 84.23 & 84.58 & 90.00 & 85.51 & 79.68 & 81.57 & 81.19 & 77.88 \\
B & SoTTA* & 71.06 & 80.72 & 73.98 & 82.02 & 67.78 & 79.96 & 83.85 & 81.16 & 81.96 & 80.95 & 87.10 & 82.77 & 74.12 & 78.02 & 75.80 & 78.29 \\
B & SimATTA* & 33.37 & 49.99 & 41.33 & 62.69 & 58.03 & 76.02 & 81.32 & 77.35 & 80.75 & 79.95 & 88.83 & 67.17 & 76.13 & 72.59 & 78.84 & 68.29 \\
\rowcolor[HTML]{DCFFDC} 
B & \method{} & \textbf{75.72} & \textbf{83.25} & \textbf{78.58} & \textbf{85.41} & \textbf{75.75} & \textbf{86.14} & \textbf{89.82} & \textbf{87.28} & \textbf{89.55} & \textbf{88.83} & \textbf{93.67} & \textbf{92.04} & \textbf{84.93} & \textbf{91.91} & \textbf{88.38} & \textbf{86.08} \\
\bottomrule
\end{tabularx}
}
\vspace{0.05cm}
\caption{ResNet50.}
\end{subtable}

\begin{subtable}{\textwidth}
\resizebox{\columnwidth}{!}{%
\begin{tabularx}{1.3\textwidth}{c l *{16}Y}
\toprule
& & \multicolumn{3}{c}{Noise} & \multicolumn{4}{c}{Blur} & \multicolumn{4}{c}{Weather} & \multicolumn{4}{c}{Digital} &  \\
& & \multicolumn{15}{l}{ %$t$ 
    \begin{tikzpicture}[overlay, remember picture]
      \draw[->, line width=0.2pt] (0,0.075) -- (19,0.075);
    \end{tikzpicture}
    } &  \\
\multirow{-3}{*}{Label} & \multirow{-3}{*}{Method} & Gau. & Shot & Imp. & Def. & Gla. & Mot. & Zoom & Snow & Fro. & Fog & Brit. & Cont. & Elas. & Pix. & JPEG & Avg. \\
\midrule
- & SrcValid & 39.84 & 41.82 & 34.89 & 55.54 & 56.59 & 58.42 & 60.38 & 65.70 & 57.74 & 43.48 & 72.40 & 20.83 & 68.84 & 66.77 & 66.92 & 54.01 \\
% - & BN-Stats & 39.84 & 41.82 & 34.89 & 55.54 & 56.59 & 58.42 & 60.38 & 65.70 & 57.74 & 43.48 & 72.40 & 20.83 & 68.84 & 66.77 & 66.92 & 54.01 \\
B & TENT* & 34.98 & 37.63 & 28.47 & 55.19 & 37.90 & 55.62 & 63.70 & 53.38 & 56.36 & 48.73 & 62.57 & 20.46 & 60.85 & 63.05 & 62.21 & 49.41 \\
B & EATA* & \textbf{55.23} & 61.37 & 51.56 & 63.45 & 62.71 & 69.75 & 74.29 & 67.57 & 68.86 & 58.67 & 72.04 & 29.22 & 68.66 & 70.93 & 69.05 & 62.89 \\
B & SAR* & 40.74 & 44.40 & 36.36 & 57.77 & 58.43 & 63.49 & 66.87 & 67.01 & 62.74 & 52.41 & 72.93 & 22.30 & 71.71 & 62.83 & 68.83 & 56.59 \\
B & CoTTA* & 39.84 & 41.83 & 34.89 & 55.54 & 56.56 & 58.43 & 60.38 & 65.68 & 57.69 & 43.46 & 72.36 & 20.82 & 68.90 & 66.77 & 66.94 & 54.01 \\
B & RoTTA* & 38.89 & 38.69 & 30.53 & 61.44 & 52.76 & 66.21 & 71.65 & 61.72 & 54.29 & 59.08 & 74.23 & 29.24 & 71.30 & 58.42 & 67.70 & 55.74 \\
B & SoTTA* & 54.59 & 62.22 & 51.71 & 63.62 & 61.67 & 67.45 & 72.72 & 66.28 & 67.09 & 58.19 & 70.80 & 30.92 & 67.18 & 70.71 & 67.34 & 62.17 \\
B & SimATTA* & 54.08 & 60.09 & 47.30 & 65.07 & 57.94 & 66.37 & 70.90 & 64.23 & 66.15 & 56.46 & 71.65 & 27.98 & 67.25 & 68.31 & 67.40 & 61.26 \\
\rowcolor[HTML]{DCFFDC} 
B & \method{} & 53.83 & \textbf{62.40} & \textbf{52.23} & \textbf{66.77} & \textbf{63.88} & \textbf{72.27} & \textbf{77.66} & \textbf{69.97} & \textbf{72.40} & \textbf{65.10} & \textbf{76.07} & \textbf{30.70} & \textbf{73.80} & \textbf{74.84} & \textbf{72.65} & \textbf{65.64} \\
\bottomrule
\end{tabularx}
}
\vspace{0.05cm}
\caption{ViT-Base.}
\end{subtable}

\end{table}

\vspace{-0.15cm}\paragraph{Comparison with original TTA and active TTA.} In Tables~\ref{tab:app_original} and \ref{tab:app_original_pacs}, we compare \method{} with \underline{original TTA (without binary-feedback samples) and original active TTA (with full-labeling)} baselines. We first observe that comparing to \system{} setting, unsupervised adaptation shows 1.14\%p accuracy degradation on average; showcasing the importance of binary-feedback for guiding model adaptation.
Also, the experiment results demonstrate the superior performance of \method{}, even outperforming the active TTA baseline (SimATTA, \cite{atta}), showing the effectiveness of our RL-based adaptation with binary-feedback adaptation and agreement-based adaptation. We consider this the drawback of SimATTA's strategy of using source-like confident samples. Even with tuning the hyperparameters, the accuracy of source-like samples is highly dependent on the source-pretrained model. This results in noisy predictions, hindering its applicability in various datasets and scenarios.

\begin{table}[t]
\centering
\caption{Accuracy (\%) and standard deviation comparisons with \textbf{original TTA and full-label active TTA baselines} in corruption datasets (severity level 5). F: Full-label feedback active TTA, B: TTA with binary feedback. Results that outperform all baselines are highlighted in \textbf{bold} font. Averaged over three random seeds. }
\label{tab:app_original}
\footnotesize

\begin{subtable}{\textwidth}
    
\resizebox{\columnwidth}{!}{%
\begin{tabularx}{1.3\textwidth}{c|l *{16}Y}
\toprule
 & & \multicolumn{3}{c}{Noise} & \multicolumn{4}{c}{Blur} & \multicolumn{4}{c}{Weather} & \multicolumn{4}{c}{Digital} &  \\
 & & \multicolumn{15}{l}{ %$t$ 
    \begin{tikzpicture}[overlay, remember picture]
      \draw[->, line width=0.2pt] (0,0.075) -- (19,0.075);
    \end{tikzpicture}
    } &  \\
\multirow{-3}{*}{Label} & {\multirow{-3}{*}{Method}} & Gau. & Shot & Imp. & Def. & Gla. & Mot. & Zoom & Snow & Fro. & Fog & Brit. & Cont. & Elas. & Pix. & JPEG & Avg. \\
\midrule
{-} & SrcValid & 25.97 & 33.19 & 24.71 & 56.73 & 52.02 & 67.37 & 64.80 & 77.97 & 67.01 & 74.14 & 91.51 & 33.90 & 76.62 & 46.38 & 73.23 & 57.23 \\
- & BN-Stats & 66.96 & 69.04 & 60.36 & 87.78 & 65.55 & 86.29 & 87.38 & 81.63 & 80.28 & 85.39 & 90.74 & 86.88 & 76.72 & 79.33 & 71.92 & 78.42 \\
- & TENT & 74.34 & 77.30 & 65.86 & 74.12 & 54.40 & 58.08 & 58.89 & 53.49 & 50.45 & 46.76 & 48.23 & 40.65 & 34.78 & 34.37 & 29.62 & 53.42 \\
- & EATA & 76.45 & 77.33 & 64.70 & 77.51 & 62.31 & 71.91 & 78.34 & 75.29 & 75.24 & 78.56 & 84.68 & 83.19 & 68.81 & 70.97 & 67.18 & 74.16 \\
- & SAR & 67.94 & 69.45 & 62.82 & \textbf{87.79} & 66.18 & 86.31 & 87.38 & 81.63 & 80.28 & 85.39 & 90.74 & 86.88 & 76.72 & 79.33 & 71.98 & 78.72 \\
- & CoTTA & 66.97 & 69.04 & 60.37 & 87.78 & 65.55 & 86.30 & 87.38 & 81.63 & 80.27 & 85.39 & 90.74 & 86.88 & 76.72 & 79.33 & 71.92 & 78.42 \\
- & RoTTA & 65.21 & 71.11 & 64.77 & 85.11 & 69.73 & 87.44 & 89.95 & 86.05 & 86.60 & 87.98 & 92.73 & 88.00 & 82.53 & 85.49 & 81.11 & 81.59 \\
- & SoTTA & 74.59 & 81.22 & 74.55 & 84.74 & 71.41 & 83.33 & 87.86 & 83.68 & 84.63 & 85.51 & 90.34 & 83.09 & 78.87 & 82.88 & 77.99 & 81.65 \\
{F} & SimATTA & 73.89 & 82.45 & 73.36 & 79.97 & 72.14 & 84.13 & 88.95 & 86.22 & 89.01 & 87.94 & 92.81 & 85.21 & 80.94 & 85.93 & 83.97 & 83.13 \\
{B} & \cellcolor[HTML]{DCFFDC}\method{} & \cellcolor[HTML]{DCFFDC}\textbf{76.78} & \cellcolor[HTML]{DCFFDC}\textbf{84.24} & \cellcolor[HTML]{DCFFDC}\textbf{78.75} & \cellcolor[HTML]{DCFFDC}{87.51} & \cellcolor[HTML]{DCFFDC}\textbf{77.39} & \cellcolor[HTML]{DCFFDC}\textbf{88.38} & \cellcolor[HTML]{DCFFDC}\textbf{91.36} & \cellcolor[HTML]{DCFFDC}\textbf{89.42} & \cellcolor[HTML]{DCFFDC}\textbf{90.72} & \cellcolor[HTML]{DCFFDC}\textbf{90.30} & \cellcolor[HTML]{DCFFDC}\textbf{94.65} & \cellcolor[HTML]{DCFFDC}\textbf{92.62} & \cellcolor[HTML]{DCFFDC}\textbf{86.15} & \cellcolor[HTML]{DCFFDC}\textbf{92.42} & \cellcolor[HTML]{DCFFDC}\textbf{87.24} & \cellcolor[HTML]{DCFFDC}\textbf{87.20} \\
\bottomrule
\end{tabularx}
}
\vspace{0.05cm}
\caption{CIFAR10-C.}
\end{subtable}

\begin{subtable}{\textwidth}
    
\resizebox{\columnwidth}{!}{%
\begin{tabularx}{1.3\textwidth}{c|l *{16}Y}
\toprule
& & \multicolumn{3}{c}{Noise} & \multicolumn{4}{c}{Blur} & \multicolumn{4}{c}{Weather} & \multicolumn{4}{c}{Digital} &  \\
& & \multicolumn{15}{l}{ %$t$ 
    \begin{tikzpicture}[overlay, remember picture]
      \draw[->, line width=0.2pt] (0,0.075) -- (19,0.075);
    \end{tikzpicture}
    } &  \\
\multirow{-3}{*}{Label} & \multirow{-3}{*}{Method} & Gau. & Shot & Imp. & Def. & Gla. & Mot. & Zoom & Snow & Fro. & Fog & Brit. & Cont. & Elas. & Pix. & JPEG & Avg. \\
\midrule
- & SrcValid & 10.63 & 12.14 & \hphantom{0}7.17 & 34.86 & 19.58 & 44.09 & 41.94 & 46.34 & 34.22 & 41.08 & 67.31 & 18.47 & 50.36 & 24.91 & 44.56 & 33.18 \\
- & BN-Stats & 39.23 & 40.75 & 34.10 & 66.14 & 42.46 & 63.57 & 64.82 & 53.81 & 53.49 & 58.15 & 68.22 & 64.48 & 53.88 & 56.63 & 45.17 & 53.66 \\
- & TENT & 49.71 & 51.12 & 38.34 & 42.40 & 24.86 & 21.51 & 17.21 & \hphantom{0}9.39 & \hphantom{0}5.84 & \hphantom{0}4.24 & \hphantom{0}3.87 & \hphantom{0}2.56 & \hphantom{0}2.74 & \hphantom{0}2.40 & \hphantom{0}2.36 & 18.57 \\
- & EATA & 10.40 & \hphantom{0}2.88 & \hphantom{0}2.81 & \hphantom{0}2.50 & \hphantom{0}2.22 & \hphantom{0}2.21 & \hphantom{0}1.99 & \hphantom{0}2.17 & \hphantom{0}1.91 & \hphantom{0}1.65 & \hphantom{0}1.53 & \hphantom{0}1.23 & \hphantom{0}1.25 & \hphantom{0}1.12 & \hphantom{0}1.05 & \hphantom{0}2.46 \\
- & SAR & 46.45 & 55.24 & 48.53 & \textbf{66.27} & 50.93 & \textbf{65.35} & \textbf{68.49} & 60.73 & 62.36 & 63.37 & 71.12 & \textbf{69.48} & 59.76 & 65.34 & 56.33 & 60.65 \\
- & CoTTA & 39.24 & 40.75 & 34.11 & 66.13 & 42.46 & 63.57 & 64.82 & 53.81 & 53.49 & 58.14 & 68.22 & 64.48 & 53.87 & 56.63 & 45.17 & 53.66 \\
- & RoTTA & 35.63 & 40.04 & 35.55 & 60.32 & 42.09 & 62.76 & 67.53 & 58.54 & 60.60 & 60.72 & \textbf{71.58} & 64.08 & 59.50 & 63.13 & 54.49 & 55.77 \\
- & SoTTA & \textbf{52.31} & 57.80 & 48.30 & 61.57 & 48.82 & 63.45 & 68.17 & 59.54 & 61.69 & 62.62 & 69.73 & 66.30 & 57.40 & 63.35 & 56.67 & 59.85 \\
F & SimATTA & 42.86 & 54.18 & 44.18 & 53.98 & 46.64 & 60.51 & 65.54 & 57.01 & 62.73 & 57.25 & 68.38 & 52.17 & 54.53 & 61.10 & 56.88 & 55.86 \\
\rowcolor[HTML]{DCFFDC} 
B & \method{} & 50.12 & \textbf{58.34} & \textbf{52.07} & {63.27} & \textbf{52.70} & 63.80 & 68.16 & \textbf{62.65} & \textbf{65.39} & \textbf{63.79} & 71.26 & {68.97} & \textbf{63.93} & \textbf{69.45} & \textbf{63.38} & \textbf{62.49}\\
\bottomrule
\end{tabularx}
}
\vspace{0.05cm}
\caption{CIFAR100-C.}
\end{subtable}

\begin{subtable}{\textwidth}
    
\resizebox{\columnwidth}{!}{%
\begin{tabularx}{1.3\textwidth}{c|l *{16}Y}
\toprule
& & \multicolumn{3}{c}{Noise} & \multicolumn{4}{c}{Blur} & \multicolumn{4}{c}{Weather} & \multicolumn{4}{c}{Digital} &  \\
& & \multicolumn{15}{l}{ %$t$ 
    \begin{tikzpicture}[overlay, remember picture]
      \draw[->, line width=0.2pt] (0,0.075) -- (19,0.075);
    \end{tikzpicture}
    } &  \\
\multirow{-3}{*}{Label} & \multirow{-3}{*}{Method} & Gau. & Shot & Imp. & Def. & Gla. & Mot. & Zoom & Snow & Fro. & Fog & Brit. & Cont. & Elas. & Pix. & JPEG & Avg. \\
\midrule
- & SrcValid & \hphantom{0}6.99 & \hphantom{0}8.93 & \hphantom{0}5.09 & 15.18 & \hphantom{0}9.65 & 26.50 & 26.33 & 29.77 & 33.64 & 12.34 & 31.80 & \hphantom{0}2.34 & 27.71 & 34.99 & 46.97 & 21.22 \\
- & BN-Stats & 31.45 & 33.28 & 23.55 & 32.33 & 22.30 & 44.30 & 45.04 & 38.89 & 42.64 & 29.97 & 46.55 & \hphantom{0}8.46 & 43.70 & 52.53 & 49.50 & 36.30 \\
- & TENT & 35.97 & 33.92 & 18.12 & \hphantom{0}8.67 & \hphantom{0}2.93 & \hphantom{0}2.84 & \hphantom{0}2.57 & \hphantom{0}2.35 & \hphantom{0}1.87 & \hphantom{0}1.86 & \hphantom{0}1.86 & \hphantom{0}1.33 & \hphantom{0}1.57 & \hphantom{0}1.63 & \hphantom{0}1.58 & \hphantom{0}7.94 \\
- & EATA & 34.53 & 36.80 & 26.46 & \textbf{36.49} & 25.69 & 47.83 & 48.33 & 41.88 & 44.98 & 35.83 & 49.62 & \hphantom{0}6.86 & 44.86 & 53.79 & 50.95 & 38.99 \\
- & SAR & 33.35 & 38.03 & 28.94 & 35.83 & \textbf{27.12} & 47.13 & 48.39 & 41.36 & 45.09 & 36.79 & 50.24 & \textbf{13.46} & 46.45 & 52.44 & 50.52 & 39.68 \\
- & CoTTA & 31.45 & 33.29 & 23.54 & 32.35 & 22.27 & 44.33 & 44.99 & 38.94 & 42.67 & 29.99 & 46.57 & \hphantom{0}8.67 & 43.74 & 52.58 & 49.45 & 36.32 \\
- & RoTTA & 31.13 & 34.94 & 25.71 & 31.74 & 25.01 & 46.18 & 47.47 & 41.40 & 45.13 & 31.38 & 48.01 & \hphantom{0}8.92 & 45.07 & 50.77 & 49.69 & 37.50 \\
- & SoTTA & \textbf{37.62} & \textbf{40.91} & \textbf{31.72} & 33.55 & 26.75 & 41.50 & 44.84 & 37.72 & 41.42 & 38.75 & 47.04 & \hphantom{0}7.46 & 34.88 & 44.08 & 45.04 & 36.89 \\
F & SimATTA & 23.70 & 33.82 & 26.11 & 23.55 & 23.36 & 40.16 & 43.41 & 30.22 & 41.84 & 26.42 & 40.72 & \hphantom{0}2.88 & 41.37 & 49.21 & {52.85} & 33.31 \\
\rowcolor[HTML]{DCFFDC} 
B & \cellcolor[HTML]{DCFFDC}\method{} & \cellcolor[HTML]{DCFFDC}34.84 & \cellcolor[HTML]{DCFFDC}{39.88} & \cellcolor[HTML]{DCFFDC}{28.56} & \cellcolor[HTML]{DCFFDC}35.37 & \cellcolor[HTML]{DCFFDC}26.65 & \cellcolor[HTML]{DCFFDC}\textbf{48.41} & \cellcolor[HTML]{DCFFDC}\textbf{49.57} & \cellcolor[HTML]{DCFFDC}\textbf{43.62} & \cellcolor[HTML]{DCFFDC}\textbf{47.90} & \cellcolor[HTML]{DCFFDC}\textbf{39.53} & \cellcolor[HTML]{DCFFDC}\textbf{50.95} & \cellcolor[HTML]{DCFFDC}12.27 & \cellcolor[HTML]{DCFFDC}\textbf{47.18} & \cellcolor[HTML]{DCFFDC}\textbf{54.01} & \cellcolor[HTML]{DCFFDC}\textbf{54.06} & \cellcolor[HTML]{DCFFDC}\textbf{40.85} \\
\bottomrule
\end{tabularx}
}
\vspace{0.05cm}
\caption{Tiny-ImageNet-C.}
\end{subtable}

\end{table}

\begin{table}[t]
\centering
\caption{Accuracy (\%) and standard deviation comparisons with \textbf{original TTA and full-label active TTA baselines} in PACS. The domain-wise data stream is a continual TTA setting~\citep{cotta}, and the mixed data stream shuffled all domains randomly, where we report the cumulative accuracy at each of the four adaptation points. F: Full-label feedback active TTA, B: TTA with binary feedback. Results outperforming all other baselines are highlighted in \textbf{bold} fonts. Averaged over three random seeds.}

\label{tab:app_original_pacs}
\scriptsize	
\begin{tabularx}{\textwidth}{c|l *{8}Y}

\toprule

& & \multicolumn{4}{c}{Domain-wise data stream} & \multicolumn{4}{c}{Mixed data stream} \\ % \cmidrule(lr){7-7}\cmidrule(lr){12-12}
& & \multicolumn{4}{l}{ %$t$ 
    \begin{tikzpicture}[overlay, remember picture]
      \draw[->, line width=0.2pt] (0,0.075) -- (7,0.075);
    \end{tikzpicture}
    } & 
    \multicolumn{4}{l}{ %$t$ 
    \begin{tikzpicture}[overlay, remember picture]
      \draw[->, line width=0.2pt] (0,0.075) -- (7,0.075);
    \end{tikzpicture}
    } \\
{\multirow{-3}{*}{Label}} & {\multirow{-3}{*}{Method}} & Art & Cartoo- & Sketch & Avg & 25\% & 50\% & 75\% & 100\%(Avg) \\
\midrule
- & SrcValid & 59.38 \scalebox{\std}{±0.00} & 27.94 \scalebox{\std}{±0.21} & 42.96 \scalebox{\std}{±0.01} & 43.43 \scalebox{\std}{±0.07} & 42.74 \scalebox{\std}{±1.13} & 42.80 \scalebox{\std}{±0.22} & 42.64 \scalebox{\std}{±0.30} & 42.77 \scalebox{\std}{±0.01} \\
- & BN Stats & 67.87 \scalebox{\std}{±0.18} & 63.48 \scalebox{\std}{±0.88} & 54.07 \scalebox{\std}{±0.36} & 61.81 \scalebox{\std}{±0.18} & 59.09 \scalebox{\std}{±0.29} & 58.28 \scalebox{\std}{±0.08} & 58.05 \scalebox{\std}{±0.22} & 57.82 \scalebox{\std}{±0.20} \\
- & TENT & 71.61 \scalebox{\std}{±0.70} & 67.00 \scalebox{\std}{±0.51} & 44.14 \scalebox{\std}{±0.85} & 60.92 \scalebox{\std}{±0.29} & {\ul 60.34 \scalebox{\std}{±0.51}} & 56.75 \scalebox{\std}{±0.62} & 53.22 \scalebox{\std}{±0.57} & 49.64 \scalebox{\std}{±0.50} \\
- & EATA & 68.44 \scalebox{\std}{±0.31} & 64.90 \scalebox{\std}{±0.69} & 58.58 \scalebox{\std}{±0.18} & 63.97 \scalebox{\std}{±0.23} & 59.60 \scalebox{\std}{±0.15} & 58.98 \scalebox{\std}{±0.54} & 59.10 \scalebox{\std}{±0.38} & 59.24 \scalebox{\std}{±0.08} \\
- & SAR & 67.90 \scalebox{\std}{±0.20} & 63.60 \scalebox{\std}{±0.83} & 55.23 \scalebox{\std}{±0.44} & 62.25 \scalebox{\std}{±0.11} & 59.13 \scalebox{\std}{±0.21} & 58.49 \scalebox{\std}{±0.15} & 58.32 \scalebox{\std}{±0.05} & 58.25 \scalebox{\std}{±0.07} \\
- & CoTTA & 67.87 \scalebox{\std}{±0.18} & 63.48 \scalebox{\std}{±0.88} & 54.06 \scalebox{\std}{±0.35} & 61.81 \scalebox{\std}{±0.19} & 59.10 \scalebox{\std}{±0.32} & 58.29 \scalebox{\std}{±0.09} & 58.06 \scalebox{\std}{±0.23} & 57.83 \scalebox{\std}{±0.22} \\
- & RoTTA & 64.39 \scalebox{\std}{±0.59} & 38.27 \scalebox{\std}{±0.61} & 40.80 \scalebox{\std}{±1.64} & 47.82 \scalebox{\std}{±0.20} & 52.64 \scalebox{\std}{±0.25} & 49.01 \scalebox{\std}{±0.85} & 46.87 \scalebox{\std}{±0.55} & 45.75 \scalebox{\std}{±0.49} \\
- & SoTTA & 69.86 \scalebox{\std}{±0.78} & 32.02 \scalebox{\std}{±1.52} & 23.66 \scalebox{\std}{±1.77} & 41.84 \scalebox{\std}{±0.34} & 51.96 \scalebox{\std}{±5.47} & 49.84 \scalebox{\std}{±6.14} & 48.09 \scalebox{\std}{±6.64} & 47.06 \scalebox{\std}{±6.03} \\
F & SimATTA & \textbf{77.13 \scalebox{\std}{±0.76}} & 71.46 \scalebox{\std}{±2.47} & \textbf{78.80 \scalebox{\std}{±0.53}} & \textbf{75.80 \scalebox{\std}{±0.74}} & \textbf{68.27 \scalebox{\std}{±1.24}} & \textbf{72.67 \scalebox{\std}{±0.45}} & \textbf{75.41 \scalebox{\std}{±0.30}} & \textbf{77.47 \scalebox{\std}{±0.44}} \\
\rowcolor[HTML]{DCFFDC} 
B & \method{} & {\ul 73.86 \scalebox{0.65}{±3.76}} & \textbf{76.81 \scalebox{0.65}{±2.45}} & {\ul 76.03 \scalebox{0.65}{±1.61}} & {\ul 75.57 \scalebox{0.65}{±0.93}} & \cellcolor[HTML]{DCFFDC}59.65 \scalebox{0.65}{±0.70} & \cellcolor[HTML]{DCFFDC}{\ul 64.70 \scalebox{0.65}{±0.78}} & \cellcolor[HTML]{DCFFDC}{\ul 69.23 \scalebox{0.65}{±0.17}} & \cellcolor[HTML]{DCFFDC}{\ul 72.18 \scalebox{0.65}{±0.38}}\\
\bottomrule

\end{tabularx}
\end{table}

\vspace{-0.15cm}\paragraph{Comparison with enhanced TTA.}
Following the setting of SimATTA~\citep{atta}, we compare \method{} with an enhanced TTA setting, which is \textbf{unsupervised TTA baselines adapting on the fine-tuned model}, which is tuned with an equal amount of binary-feedback samples before the adaptation phase. In Table~\ref{tab:app_enhanced_corruption}, we observe that \method{} still outperforms over enhanced TTA baselines. The result necessitates the superiority of online adaptation on binary feedback samples.

\begin{table}[t]
\centering
\caption{Accuracy (\%) comparisons with \textbf{enhanced TTA on fine-tuned model}~\cite{atta} and TTA with binary feedback baselines on source model, in CIFAR10-C (severity level 5). Notation * indicates the modified algorithm to utilize binary-feedback samples. E: Enhanced TTA, B: TTA with binary feedback. Results outperforming all other baselines are highlighted in \textbf{bold} fonts. Averaged over three random seeds.}
\label{tab:app_enhanced_corruption}
\footnotesize

\resizebox{\columnwidth}{!}{%
\begin{tabularx}{1.3\textwidth}{c|l  *{16}Y}
\toprule
 & & \multicolumn{3}{c}{Noise} & \multicolumn{4}{c}{Blur} & \multicolumn{4}{c}{Weather} & \multicolumn{4}{c}{Digital} &  \\
 & & \multicolumn{15}{l}{ %$t$ 
    \begin{tikzpicture}[overlay, remember picture]
      \draw[->, line width=0.2pt] (0,0.075) -- (19,0.075);
    \end{tikzpicture}
    } &  \\
{\multirow{-3}{*}{Label}} & {\multirow{-3}{*}{Method}} & Gau. & Shot & Imp. & Def. & Gla. & Mot. & Zoom & Snow & Fro. & Fog & Brit. & Cont. & Elas. & Pix. & JPEG & Avg. \\
\midrule
E & SrcValid & 76.17 & 77.48 & 67.54 & 82.24 & 71.89 & 79.90 & 83.44 & 82.67 & 84.36 & 81.18 & 88.74 & 75.12 & 77.53 & 80.66 & 80.24 & 79.28 \\
E & BN-Stats & 77.90 & 79.66 & 71.76 & 86.52 & 73.53 & 85.26 & 86.77 & 84.66 & 85.27 & 84.07 & 90.10 & 86.70 & 79.39 & 84.76 & 78.98 & 82.36 \\
E & TENT & 77.52 & 76.94 & 63.79 & 68.35 & 52.67 & 56.00 & 55.58 & 52.93 & 49.02 & 45.02 & 43.94 & 33.46 & 32.12 & 31.39 & 29.27 & 51.20 \\
E & EATA & 77.18 & 75.32 & 64.66 & 70.73 & 58.46 & 64.62 & 70.22 & 68.00 & 68.34 & 67.35 & 75.81 & 69.52 & 62.93 & 69.02 & 64.28 & 68.43 \\
E & SAR & 77.90 & 79.66 & 71.76 & 86.52 & 73.53 & 85.26 & 86.77 & 84.66 & 85.27 & 84.07 & 90.10 & 86.70 & 79.39 & 84.76 & 78.98 & 82.36 \\
E & CoTTA & 77.90 & 79.66 & 71.77 & 86.52 & 73.53 & 85.26 & 86.77 & 84.66 & 85.27 & 84.06 & 90.09 & 86.71 & 79.39 & 84.76 & 78.98 & 82.36 \\
E & RoTTA & 78.93 & 81.00 & 74.28 & 86.56 & 75.45 & 86.18 & 88.63 & 86.85 & 87.71 & 86.73 & 91.36 & 88.06 & 82.41 & 87.19 & 82.42 & 84.25 \\
E & SoTTA & \textbf{79.19} & 81.45 & 74.23 & 82.67 & 70.73 & 81.99 & 85.41 & 82.78 & 83.69 & 85.02 & 89.40 & 84.41 & 78.41 & 83.44 & 78.94 & 81.45 \\
\midrule
B & SimATTA* & 70.21 & 81.67 & 71.49 & 79.59 & 69.41 & 82.15 & 87.28 & 83.90 & 86.89 & 86.49 & 91.51 & 83.40 & 77.94 & 83.81 & 82.25 & 81.09 \\
B & \cellcolor[HTML]{DCFFDC}\method{} & \cellcolor[HTML]{DCFFDC}76.78 & \cellcolor[HTML]{DCFFDC}\textbf{84.24} & \cellcolor[HTML]{DCFFDC}\textbf{78.75} & \cellcolor[HTML]{DCFFDC}\textbf{87.51} & \cellcolor[HTML]{DCFFDC}\textbf{77.39} & \cellcolor[HTML]{DCFFDC}\textbf{88.38} & \cellcolor[HTML]{DCFFDC}\textbf{91.36} & \cellcolor[HTML]{DCFFDC}\textbf{89.42} & \cellcolor[HTML]{DCFFDC}\textbf{90.72} & \cellcolor[HTML]{DCFFDC}\textbf{90.30} & \cellcolor[HTML]{DCFFDC}\textbf{94.65} & \cellcolor[HTML]{DCFFDC}\textbf{92.62} & \cellcolor[HTML]{DCFFDC}\textbf{86.15} & \cellcolor[HTML]{DCFFDC}\textbf{92.42} & \cellcolor[HTML]{DCFFDC}\textbf{87.24} & \cellcolor[HTML]{DCFFDC}\textbf{87.20}
\\
\bottomrule
\end{tabularx}
}

\end{table}

\section{Experiment details}\label{app:exp_detail}

We conducted all experiments with three random seeds [0, 1, 2] and reported the mean and standard deviation values. The experiments were mainly conducted on NVIDIA RTX 3090 and TITAN GPUs.

\subsection{Settings}

\paragraph{Dataset.} We utilized the corruption dataset (CIFAR10-C, CIFAR100-C, Tiny-ImageNet-C~\citep{cifarc}) and domain generalization baselines (PACS~\citep{pacs}). CIFAR10-C/CIFAR100-C/Tiny-ImageNet-C is a 10/100/200-class dataset of a total of 150,000 images in 15 types of image corruptions, including Gaussian, Snow, Frost, Fog, Brightness, Contrast, Elastic Transformation, Pixelate, and JPEG Compression. PACS is a 7-class dataset with 9,991 images in four domains of art painting, cartoon, photo, and sketch. 

\vspace{-0.15cm}\paragraph{Source domain pre-training.} We closely followed the settings and utilized the pre-trained weights provided by 
SoTTA~\citep{sotta} and SimATTA~\citep{atta}. As the backbone model, we employ the ResNet18~\citep{resnet} from TorchVision~\citep{torchvision2016}.
% During training, we fixed the statistics in the batch normalization layer to those of the original pre-trained model.
For CIFAR10-C/CIFAR100-C/Tiny-ImageNet-C, we trained the model with the source data with a learning rate of 0.1/0.1/0.001 and a momentum of 0.9, with cosine annealing learning rate scheduling for 200 epochs.
% For Tiny-ImageNet-C, we followed SimATTA~\citep{atta}. We adapted the pre-trained weights from ImageNet to the brightness corruption domain by training only the last linear layer for 10,000 iterations using the Adam optimizer with a learning rate of 0.001.
For PACS, we fine-tuned the pre-trained weights from ImageNet on the selected source domains for 3,000 iterations using the Adam optimizer with a learning rate of 0.0001.

\vspace{-0.15cm}\paragraph{Scenario.}
% \tk{@Sorn: please add pre-training and fine-tuning details}
For the number of binary-feedback samples, we used $k = 3$ samples from a 64-sample test batch, accounting for less than 5\% of the total data size. For the binary version of TTA baselines, we added cross-entropy loss (for correct samples) combined with complementary loss (for incorrect samples, \cite{kim2019nlnl}), maintaining an equal budget size to our method. 
\
To implement, we replace the original TTA loss $l_{\tt TTA}$ with $l_{\tt TTA} + l_{\tt CE} + l_{\tt CCE}$, where $l_{\tt CE}$ is a cross-entropy loss on correct samples and $l_{\tt CCE} = - \sum^{\tt num\_class}_{k=1} y_k \log (1 - f_\theta (k | x))$ is the complementary cross-entropy loss~\citep{kim2019nlnl} on incorrect samples.

For enhanced TTA, we used the same binary version loss with an SGD optimizer with a learning rate of 0.001 and a batch size of 64. The number of fine-tuning epochs was set to 150 for PACS, 150 for CIFAR-10, 150 for CIFAR-100, and 25 for Tiny-ImageNet-C. Note that the hyperparameters were selected to maximize accuracy on the test data stream, which is unrealistic since test data stream accuracy is not accessible during the fine-tuning process. % Therefore, the results of fine-tuned TTA presented in the paper can be considered as the upper bound of fine-tuned TTA performance in a real-world scenario.

\vspace{-0.15cm}\paragraph{Comparison with active TTA.}~\label{app:compare_active}
To compare \method{} with full-label feedback methods, we propose two scenarios: (1) an equal labeling cost and (2) an equal number of active samples. To compare with an equal labeling cost, we formulate the labeling cost with Shannon information gain~\citep{mackay2003information} as $\log (p^{-1})$ where $p$ is the probability of selecting a label. We assume the probability of each feedback strategy as $p=2^{-1}$ (correct/incorrect) and $p={\tt num\_class}^{-1}$ (select in the entire class set). The final labeling cost for binary feedback is $1$ for binary feedback and $\log ({\tt num\_class})$ for full-label feedback. Therefore, we utilize $\log ({\tt num\_class})$ times more feedback samples for \system{} setting compared to active TTA.

\subsection{TTA Baselines}

\paragraph{TENT.} For TENT~\citep{tent}, we utilize an Adam optimizer~\citep{kingma:adam} with a learning rate $LR = 0.001$, aligning with the guidelines outlined in the original paper and active TTA paper~\citep{atta}. The implementation followed the official code.\footnote{\url{https://github.com/DequanWang/tent}}

\vspace{-0.15cm}\paragraph{EATA.} For EATA~\citep{eata}, we followed the original configuration of $LR = 0.001$, entropy constant $E_0 = 0.4 \times \ln C$, where $C$ represents the number of classes. Additionally, we set the cosine sample similarity threshold $\epsilon = 0.5$, trade-off parameter $\beta = 2,000$, and moving average factor $\alpha = 0.1$. The Fisher importance calculation involved 2,000 samples, as recommended. The implementation followed the official code.\footnote{\url{https://github.com/mr-eggplant/EATA}}

\vspace{-0.15cm}\paragraph{SAR.} For SAR~\citep{sar}, we set a learning rate of $LR=0.00025$, sharpness threshold $\rho = 0.5$, and entropy threshold $E_0 = 0.4 \times \mathrm{ln} C$, following the recommendations from the original paper. The top layer (layer 4 for ResNet18) was frozen, consistent with the original paper. The implementation followed the official code.\footnote{\url{https://github.com/mr-eggplant/SAR}}

\vspace{-0.15cm}\paragraph{CoTTA.} For CoTTA~\citep{cotta}, we set the restoration factor $p=0.01$, and exponential moving average (EMA) factor $\alpha=0.999$. For augmentation confidence threshold $p_{th}$, we followed the previous implementation~\citep{atta} as $p_{th}=0.1$. The implementation followed the official code.\footnote{\url{https://github.com/qinenergy/cotta}}

\vspace{-0.15cm}\paragraph{RoTTA.} For RoTTA~\citep{rotta}, we utilized the Adam optimizer~\citep{kingma:adam} with a learning rate of $LR = 0.001$ and $\beta = 0.9$. We followed the original hyperparameters, including BN-statistic exponential moving average updating rate $\alpha = 0.05$, Teacher model's exponential moving average updating rate $\nu = 0.001$, timeliness parameter $\lambda_t = 1.0$, and uncertainty parameter $\lambda_u = 1.0$. The implementation followed the original code.\footnote{\url{https://github.com/BIT-DA/RoTTA}}

\vspace{-0.15cm}\paragraph{SoTTA.} For SoTTA~\citep{sotta}, we utilized the Adam optimizer~\citep{kingma:adam}, with a BN momentum of $m = 0.2$ and a learning rate of $LR = 0.001$. The memory size was set to 64, with the confidence threshold $C_0=0.99$. The entropy-sharpness L2-norm constraint $\rho$ was set to 0.5, aligning with the suggestion~\citep{sam}. The top layer was frozen following the original paper. 
The implementation followed the original code.\footnote{\url{https://github.com/taeckyung/sotta}}

\vspace{-0.15cm}\paragraph{SimATTA.} We follow the original implementation of SimATTA~\citep{atta}. Since SimATTA queries active samples at a dynamic rate, we set the centroid increase number to $k=3$ and limit the budget per batch to 3, ensuring an equal active sample budget compared to \method{}. For the adaptation objective, we add the complementary loss (incorrect samples, \cite{kim2019nlnl}) to the original cross-entropy loss for correct samples.
For CIFAR-10 and CIFAR-100, 
% which were not tested by the original authors of SimATTA, 
we performed a grid search to find the optimal hyperparameters. 
% Specifically, we tested SimATTA using learning rates $LR = 0.1/0.01/0.001/0.0001/0.00001$, higher entropy thresholds $e_h = 0.1/0.01/0.001/0.0001/0.00001/0.000001$, and lower entropy thresholds $e_l = 0.1/0.01/0.001/0.0001/0.00001/0.000001$. 
We found the optimal hyperparameters to be $LR = 0.0001/0.0001$, $e_h = 0.001/0.001$, and $e_l = 0.0001/0.00001$ for the CIFAR-10 and CIFAR-100 datasets, respectively. The implementation is based on the original code.\footnote{\url{https://github.com/divelab/ATTA}}

\vspace{-0.15cm}\paragraph{BiTTA.} 
During adaptation, we update all parameters, including BN stats, with an SGD optimizer with a learning rate/epoch of 0.001/3 (PACS), 0.0001/3 (CIFAR10-C, CIFAR100-C), and 0.00005/5 (Tiny-ImageNet-C) on the entire model. We applied weight decay of 0.05 to PACS and 0.0 otherwise. We applied stochastic restoration~\citep{cotta} in Tiny-ImageNet-C to prevent overfitting. We update batch norm statistics with momentum 0.3 on the unlabeled test batch, and freeze the statistics during adaptation, following \citet{atta}. We apply the dropout layer after residual blocks, following the previous work on TTA accuracy estimation~\citep{aetta}. With 4 dropout instances, we apply a dropout rate of 0.3 for small-scale datasets (e.g., CIFAR10-C, CIFAR100-C, PACS) and 0.1 for large-scale datasets (e.g., Tiny-ImageNet-C, ImageNet-C). 

\section{License of assets}\label{app:license}

\paragraph{Datasets.} CIFAR10-C/CIFAR100-C (Creative Commons Attribution 4.0 International), and Tiny-ImageNet-C dataset (Apache-2.0). The license of PACS dataset is not specified. 

\vspace{-0.15cm}\paragraph{Codes.} Torchvision for ResNet18 (Apache 2.0), the official repository of TENT (MIT License), the official repository of EATA (MIT License), the official repository of SAR (BSD 3-Clause License), the official repository of CoTTA (MIT License), the official repository of RoTTA (MIT License), the official repository of SoTTA (MIT License), and the official repository of SimATTA (GPL-3.0 License).
%, the official repository of EADA (MIT License), and the official repository of CLUE (MIT License). The licenses of ELPT, HTPL, and DiaNA are not specified.

%%%%%%%%%%%%%%%%%%%%%%%%%%%%%%%%%%%%%%%%%%%%%%%%%%%%%%%%%%%%%%%%%%%%%%%%%%%%%%%
%%%%%%%%%%%%%%%%%%%%%%%%%%%%%%%%%%%%%%%%%%%%%%%%%%%%%%%%%%%%%%%%%%%%%%%%%%%%%%%

\end{document}